\documentclass{article}

\usepackage{microtype}
\usepackage{graphicx}
\usepackage{subcaption}
\usepackage{booktabs} % for professional tables
\usepackage{adjustbox}          
\usepackage[table]{xcolor}      
\usepackage{colortbl}             
\usepackage{booktabs}         
\usepackage{array}
\usepackage{multirow} 
\usepackage{tcolorbox}
\tcbuselibrary{breakable, skins}

\usepackage{hyperref}

\usepackage[accepted]{icml2026}

\usepackage{amsmath}
\usepackage{amssymb}
\usepackage{mathtools}
\usepackage{amsthm}

\newtheorem{problem}{Problem}

\usepackage[capitalize,noabbrev]{cleveref}

\theoremstyle{plain}

\theoremstyle{definition}

\theoremstyle{remark}

\usepackage[textsize=tiny]{todonotes}

\icmltitlerunning{Diagnosing Multi-step Reasoning Failures in Black-box LLMs via Stepwise Confidence Attribution}

\begin{document}

\twocolumn[
  \icmltitle{Diagnosing Multi-step Reasoning Failures in Black-box LLMs \\ via
Stepwise Confidence Attribution}

  % Where Does Multi-step Reasoning Fail? \\ Stepwise Confidence Attribution for Closed-source LLMs}

%Where Multi-step Reasoning Fails: Step-wise Confidence Attribution in Black-box LLMs



  \icmlsetsymbol{equal}{*}

  \begin{icmlauthorlist}
    \icmlauthor{Xiaoou Liu}{ASU}
    \icmlauthor{Tiejin Chen}{ASU}
    \icmlauthor{Dengjia Zhang}{JHU}
    \icmlauthor{Yaqing Wang}{Purdue}
    \icmlauthor{Lu Cheng}{UIC}
    \icmlauthor{Hua Wei}{ASU}
  \end{icmlauthorlist}

  \icmlaffiliation{ASU}{Arizona State University}
  \icmlaffiliation{JHU}{Johns Hopkins University}
  \icmlaffiliation{Purdue}{Purdue University}
  \icmlaffiliation{UIC}{University of Illinois Chicago}
  \icmlcorrespondingauthor{Hua Wei}{hua.wei@asu.edu}

  \icmlkeywords{Machine Learning, ICML}

  \vskip 0.3in
]



\printAffiliationsAndNotice{}  % no special notice (required even if empty)

\begin{abstract}

Large Language Models have achieved strong performance on reasoning tasks with objective answers by generating step-by-step solutions, but diagnosing where a multi-step reasoning trace might fail remains difficult. Confidence estimation offers a diagnostic signal, yet existing methods are restricted to final answers or require internal model access. 
We introduce Stepwise Confidence Attribution (SCA), a framework for closed-source LLMs that assigns step-level confidence based only on generated reasoning traces. SCA applies the Information Bottleneck principle: steps aligning with consensus structures across correct solutions receive high confidence, while deviations are flagged as potentially erroneous.
We propose two complementary methods: (1) NIBS, a non-parametric IB approach measuring consistency without graph structures, and (2) GIBS, a graph-based IB model that learns subgraphs through a differentiable mask to capture logical variability. Extensive experiments on mathematical reasoning and multi-hop question answering show that SCA reliably identifies low-confidence steps strongly correlated with reasoning errors. Moreover, using step-level confidence to guide self-correction improves the correction success rate by up to 13.5\% over answer-level feedback.
%Moreover, incorporating step-level confidence into self-correction improves accuracy by up to 12.3\%. 

\end{abstract}

\section{Introduction}
\label{sec:intro}
%\vspace{-3mm}

The ability to diagnose where a multi-step reasoning trace fails is essential for improving the reliability of large language models (LLMs). Solution traces such as Chain-of-Thought~(CoT)~\citep{wei2022chain} or Graph-of-Thought~(GoT)~\citep{besta2024graph} provide transparency into model reasoning, yet intermediate errors remain hard to identify and can critically affect the final prediction. 
Recent work has explored step-level diagnostics for reasoning traces, largely falling into two categories. The first trains supervised classifiers with step-by-step human annotations to label whether each reasoning step is correct~\citep{jiao2025trustworthy, zheng2024processbench,lightman2023prm800k}. The second prompts the LLM itself as a judge to critique each solution step by step~\citep{weng2023large, li2024llm-as-judge}. While both directions can provide useful signals, the former requires expensive human annotation, and the latter inherits bias and inconsistency from the judge model~\cite{chen2026position}, limiting scalability and reliability.

\begin{figure*}
    \centering
    \includegraphics[width=0.8\linewidth]{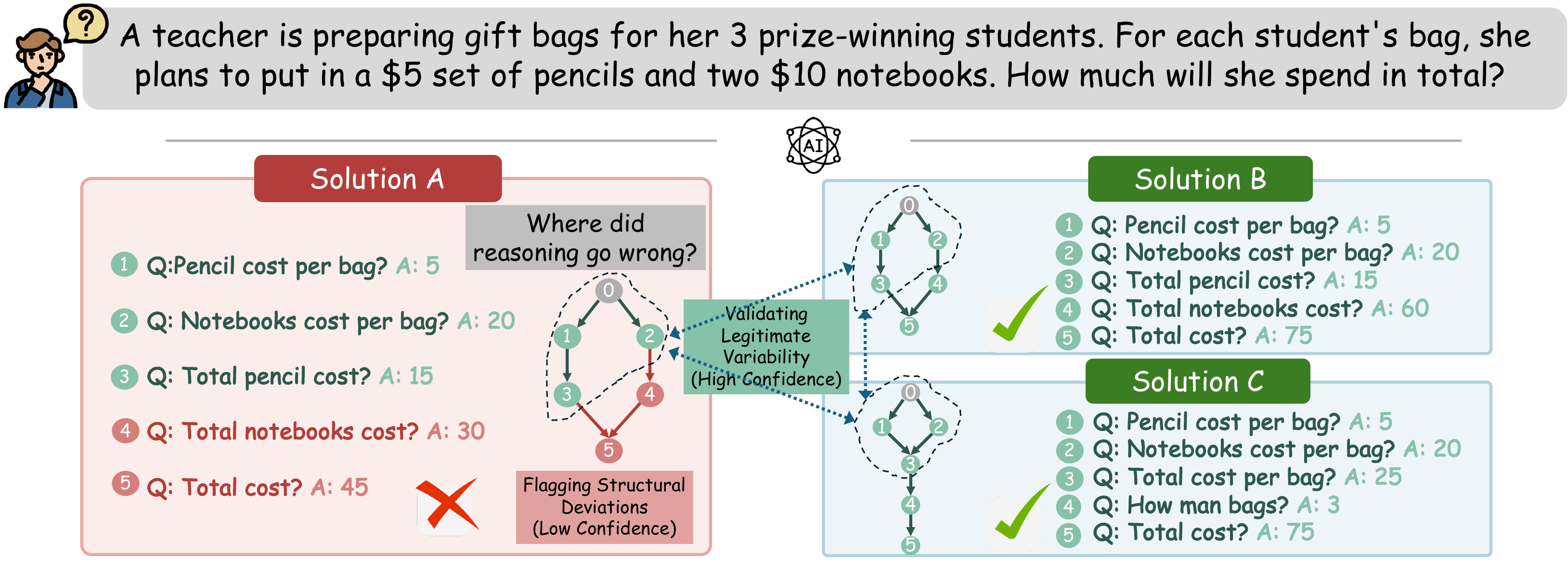}
    \vspace{-2mm}
    \caption{\small Example of reasoning trace variability in GSM8K dataset. Two distinct solution paths (B and C) yield the same correct answer, while another path (A) contains an erroneous step leading to a wrong result. Stepwise confidence attribution needs to distinguish legitimate variability from true logical inconsistencies. }
    \label{fig:intro_example}
    \vspace{-4mm}
\end{figure*}

Confidence estimation (CE) provides a complementary direction for assessing reliability, as it can operate directly on model outputs~\cite{liu2025uncertainty}. Prior work has shown that measures such as semantic variance across sampled generations~\citep{lingenerating,golovnevaroscoe} or predictive entropy over logits~\citep{lin2024contextualized, kuhnsemantic} provide informative signals for estimating whether a final answer is correct. However, restricting CE to the final answer yields only a coarse reliability signal and fails to indicate which specific step in a reasoning trace is responsible for an error. This limitation motivates the need for Stepwise Confidence Attribution (SCA), where the goal is to assign confidence scores to individual reasoning steps and thereby provide fine-grained diagnostic signals.

Extending confidence attribution from the final-answer level to the step-wise setting introduces a key challenge. Reasoning traces generated by LLMs exhibit substantial \textit{output variability}: correct solutions may differ in step order, expression, or level of detail. Yet, despite these surface-level differences, correct reasoning paths are not random; they are governed by the underlying logic of the problem. Consequently, valid traces tend to converge on specific \textit{key logical invariants}, i.e, critical intermediate states or semantic milestones that are necessary to derive the correct answer (e.g., calculating intermediate costs in a math problem). 
A robust SCA must distinguish this legitimate variability from true errors. While legitimate variability represents alternative trajectories between these logical anchors, errors act as deviations that drift away from the consensus of valid reasoning.
For example, in~\cref{fig:intro_example}, Solutions B and C arrive at the correct answer through different computation orders. Despite the permutation of steps, both solutions explicitly resolve the necessary intermediate values (the cost of pencils and notebooks). They share functional equivalence in their reasoning logic and should receive high confidence. In contrast, solution A, while structurally similar to B, introduces an incorrect operation at step 4; this deviation undermines the common structure and should receive low confidence.

Our key idea is to aggregate multiple solution traces across correct solutions and identify logical invariants. These logical invariants serve as anchors of reliable reasoning and are assigned high confidence, while steps absent from consensus patterns are assigned low confidence, since they are more likely to reflect spurious or error-prone reasoning. Consensus patterns thus act as a proxy for the latent logical pattern underlying a problem, enabling fine-grained and robust confidence attribution.

This intuition of finding a shared pattern amidst noisy variations maps naturally onto the \textbf{Information Bottleneck (IB)} principle. IB provides a formal language for balancing two competing objectives: compressing the input trajectory by discarding non-essential variations (the compression term), while retaining maximal information about the underlying correct reasoning pattern (the relevance term).
Here, the input $X$ is a reasoning trajectory composed of multiple steps, the compressed representation $Z$ is the trajectory expressed with confidence weights on its steps, and the target $Y$ denotes the correctness signal of the trajectory. 
The objective is $
\min_{Z} \; I(X;Z) - \beta I(Z;Y),$
where the first term encourages compression by selecting only a sparse subset of steps, and the second term ensures that this subset retains critical information required for correctness. 
%and the second term ensures that the retained steps remain predictive of correctness.  
%
Within this framework, we explore two complementary instantiations:

\vspace{-3mm}
\begin{itemize}
    \item \textbf{Non-parametric IB for Stepwise Confidence.} 
    $Z$ is realized as a set of consensus steps derived from correct solutions, and step-level confidence is obtained by measuring how well a trajectory aligns with this set.
    
\vspace{-1mm}
    \item \textbf{Graph IB for Stepwise Confidence (GIBS).} 
    To better handle structural variability, trajectories are represented as graphs, and $Z$ is a subgraph selected through a differentiable mask. 
    Confidence scores are produced by aligning the selected subgraph with correctness signals, providing a more flexible treatment.
\end{itemize}

\vspace{-2mm}
Empirical results show that both IB-based methods produce accurate confidence estimates. Beyond diagnostics, we demonstrate the impact of fine-grained CE on downstream LLM performance, showing that using step-level signals guides self-correction and improves correction success rate by up to 13.5\% over answer-level feedback. Ablation studies verify the contribution of each component, and robustness analyses demonstrate label availability conditions, generalization across reasoning formats, and domain shift.

\section{Related Work}

\paragraph{Confidence Estimation in LLMs.}
% Uncertainty quantification (UQ) captures distribution-level variability~\citep{gal2016dropout, malinin2018predictive} while 
Confidence estimation (CE) aims at assigning reliability scores to individual outputs~\citep{liu2025uncertainty}. 
Most CE methods focus on the final answer, either through internal signals such as entropy~\citep{kuhnsemantic,lin2024contextualized,patel2026llm}, or through black-box signals such as agreement across sampled outputs~\citep{lingenerating,chen2025uncertainty,da2025understanding,chen2026every}. 
However, these methods give no insight into intermediate steps. 
Recent stepwise CE approaches~\citep{yeuncertainty,han2025mind} attempt to assign confidence along reasoning traces, but require internal access to token probabilities, restricting them to open-source models. 
Some works also model reasoning as graphs~\citep{besta2024graph,pandey2025adaptive}, but the representation itself is orthogonal to our problem setting. 
Our work differs by introducing the problem of stepwise confidence attribution in the black-box setting, which provides scalable confidence signals at the step level using only generated reasoning traces. 

\vspace{-3mm}

\paragraph{Reasoning Verification in LLMs.}
Reasoning verification studies whether multi-step reasoning is correct. Reasoning verification is categorized into answer-level and step-level approaches. Answer-level methods, such as self-consistency~\citep{wang2022self}, outcome-based verifiers~\citep{uesato2022solving,zhang2024generative}, and LLM judges~\citep{li2024llm-as-judge}, often lack process diagnosability~\citep{tyen2023llms}. Step-level methods assess intermediate steps via human supervision~\citep{lightman2023prm800k,zheng2024processbench}, automated rewards~\citep{wang2023math,setlur2024rewarding}, or graph structures~\citep{cao2023graphreason,fang2025graph,mukherjeeparc}. However, these rely on costly annotations~\citep{lightman2023prm800k} or unreliable subjective judgments~\citep{szymanski2024limitationsllmasajudgeapproachevaluating,stechly2024selfverificationlimitationslargelanguage,jacovi2024chain}. We address this by introducing quantitative confidence signals to reduce cost and bias.

\section{Problem Statement}

Standard Answer-level Confidence Estimation (CE) assigns a reliability score to the final output $A$. (For a formal definition of Answer-level CE and distinctions between white-box and black-box settings, please refer to Appendix~\ref{app:problem_formulation}.)
While effective, answer-level CE provides only a coarse signal, failing to reveal \textit{which reasoning steps} contribute to success or failure. This limitation is critical in high-stakes decision-making where identifying the location of error is essential. 

To address this, we shift granularity to the intermediate reasoning process. For reasoning tasks, models generate a \textit{reasoning trajectory} $y_i = (T_i, A_i)$, where $T_i = \{t_{i1}, \dots, t_{iL_i}\}$ is a sequence of steps.
We introduce \textit{Stepwise Confidence Attribution (SCA)}, utilizing final-answer correctness to anchor step-level reliability.
It should be noted that in this paper, we distinguish between model certainty and step reliability. In a white-box setting, a step may be generated with high token probability (high certainty) yet be logically flawed; conversely, a correct step may have low probability if it is novel. 
Therefore, in this work, we refer to ``confidence'' not as subjective generation probability, but as a \textit{reliability score} quantifying a step's contribution to a correct answer.

\begin{problem}[Stepwise Confidence Attribution]
\label{prob:SCA}
Given an input $x$ and $N$ sampled trajectories $\mathcal{S} = \{(T_i, A_i, z_i)\}_{i=1}^N$ with correctness labels $z_i \in \{0,1\}$, the goal is to learn a mapping
$
f: (T_i, \mathcal{S}) \rightarrow \{c_{ij}\}_{j=1}^{L_i},
$
where $c_{ij}$ is the confidence score assigned to step $t_{ij}$. Here, $c_{ij}$ reflects the step's alignment with common structures across correct trajectories, serving as a proxy for correctness attribution.
\end{problem}

The primary goal of SCA is diagnostic. For an incorrect trajectory ($z_i=0$), $f$ should assign low confidence to steps responsible for the error. For a correct trajectory ($z_i=1$), steps matching the logical consensus of valid solutions should receive high confidence.
While this assumes final-answer labels, such signals are naturally available in evaluation scenarios~\citep{liu2023g,augenstein2024factuality,gao2025llm,zhao2024auto,wang2023aligning}, avoiding the need for costly step-level annotations. We also show possible adaptations for label-free settings in~\cref{sec:weak_supervision}.  
%\vspace{-2mm}
\section{Method}

Extending confidence estimation to the stepwise setting introduces the core challenge of distinguishing benign output variability from true reasoning errors. To tackle this challenge, our approach is rooted in a key insight: while individual correct solutions may vary on the surface, correct traces often share a latent common structure that reflects the essential reasoning pattern. A robust attribution method must therefore learn to identify this shared structure and assign confidence scores based on each step's consistency with it.
Our method captures these latent structures by constructing consensus anchors from correct reasoning traces. 
We leverage these consensus anchors to guide confidence attribution, formalizing the problem under the \textbf{Information Bottleneck (IB)} principle.

\subsection{Information Bottleneck Formulation}

We cast stepwise confidence attribution as an instance of the Information Bottleneck (IB) principle.  
Given a trajectory $T_i = \{t_{i1}, \dots, t_{iL_i}\}$ sampled from the LLM, the goal is to produce a confidence mask $Z = \{c_{ij}\}_{j=1}^{L_i}$ over its steps.  
Unlike settings with step-level annotations, we do not observe the correctness of individual steps.  
Instead, we only have access to final-answer labels $z_i \in \{0,1\}$ for each trajectory.  
The target $Y$ must therefore be derived from the sampled set $\mathcal{S} = \{(T_i, A_i, z_i)\}$.  
Concretely, $z_i$ partitions $\mathcal{S}$ into correct and incorrect subsets, and consensus anchors $\mathbf{m}_{ij}$ are aggregated from $\mathcal{S}_{\text{correct}}$ to approximate the latent reasoning structure.  
The IB objective is
\begin{equation}
\label{obj:ib}
    \min_{Z} \; I(T_i; Z) - \beta I(Z; Y),
\end{equation}
where $I(T_i;Z)$ encourages \textit{compression} by sparsely selecting steps, and $I(Z;Y)$ ensures \textit{relevance} between the retained steps and correctness signals inferred from consensus. In the following, we instantiate this principle with two variants: (1) \textbf{NIBS}, a non-parametric IB approach, and (2) \textbf{GIBS}, a graph-based model that applies a differentiable IB mask to handle structural variability.

\begin{figure*}
    \centering
    \includegraphics[width=0.95\linewidth]{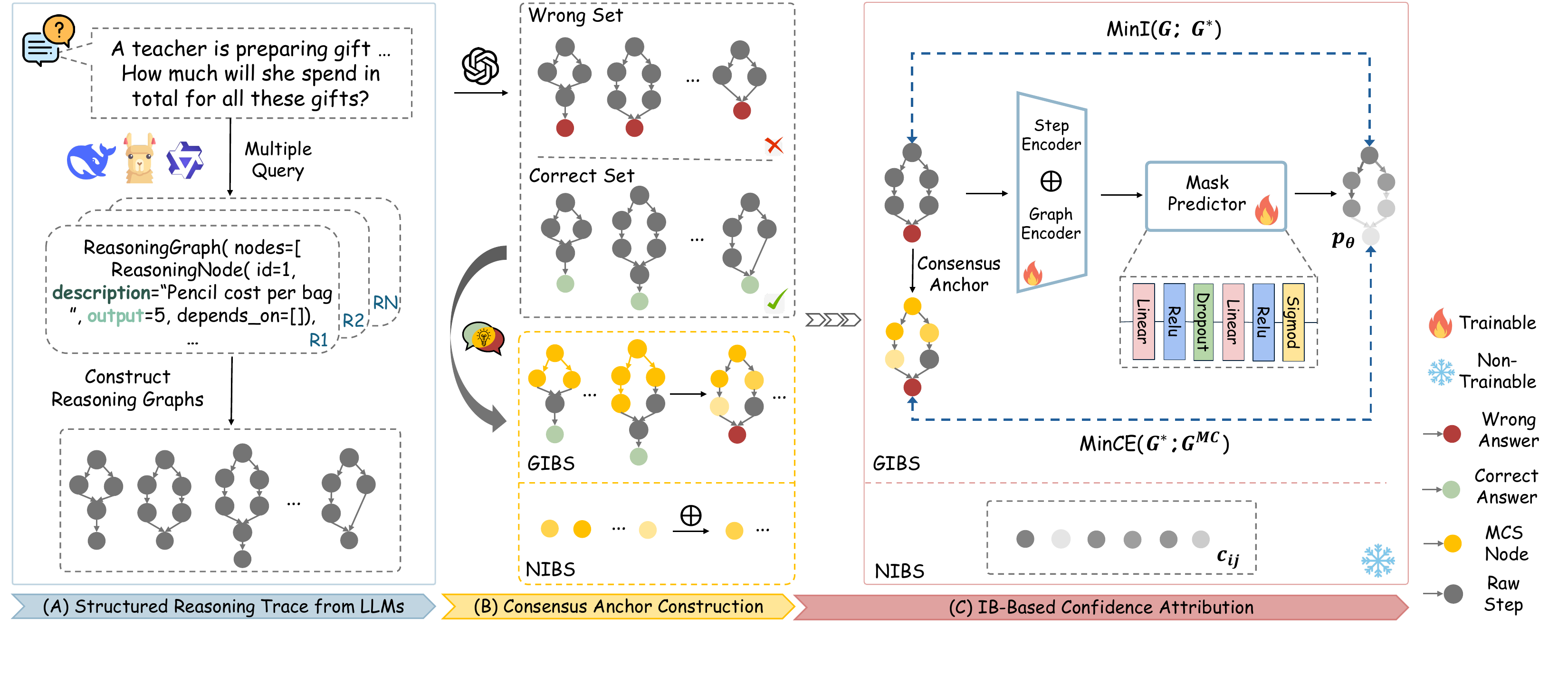}
    \vspace{-6mm}
    \caption{\small Overview of the IB-based stepwise confidence attribution framework. The process consists of (A) constructing structured reasoning traces from LLM outputs, (B) deriving consensus anchors from correct trajectories, and (C) applying the IB formulation through NIBS and GIBS to produce confidence scores.}
    \vspace{-5mm}
    \label{fig:overview}
\end{figure*}

\subsection{Non-parametric IB for Stepwise Confidence (NIBS)}

Directly solving the IB objective in Eq.~\ref{obj:ib} is generally intractable, since it requires searching over all possible confidence values of steps in $T_i$ and estimating mutual information terms. A natural approximation is to assume that steps consistently appearing in correct trajectories are the most informative about correctness $Y$, while steps absent from correct trajectories carry little predictive value.  
Under this approximation, the IB objective reduces to retaining consensus steps as the compressed representation $Z$, and assigning higher confidence to steps that align with this consensus.  
Formally, given a trajectory $T_i = \{t_{i1}, \dots, t_{iL_i}\}$ and one of its steps $t_{ij}$, its confidence can be computed as
\begin{equation}
\label{eq:sim-CE}
c_{ij} \;=\; \mathbb{E}_{S \sim \mathcal{S}_{\text{correct}}}\!\left[
\operatorname{Agg}\!\Big(\{\operatorname{sim}(\mathbf{t}_{ij}, \mathbf{t}') \;|\; \mathbf{t}' \in S\}\Big)
\right],
\end{equation}
where $\operatorname{sim}(\cdot,\cdot)$ measures semantic similarity between steps (e.g., cosine~\citep{golovnevaroscoe} or NLI~\citep{hedeberta}), and $\operatorname{Agg}$ aggregates similarities within a trajectory (e.g., maximum or mean). 

This construction instantiates the IB principle: compression $I(T_i;Z)$ is realized by restricting $Z$ to consensus steps, while relevance $I(Z;Y)$ is maximized because consensus overlap is strongly correlated with correctness.  
NIBS provides a closed-form, non-parametric, and training-free solution to the IB objective. However, NIBS ignores structural dependencies by matching steps based on the semantic similarity alone, potentially grouping nodes that are structurally disparate. This limitation motivates a graph-based IB formulation with learned subgraph selection.

%\vspace{-2mm}

\subsection{Graph IB for Stepwise Confidence (GIBS)}
%\vspace{-2mm}

While NIBS provides a simple closed-form instantiation of the IB principle, it ignores structural dependencies among reasoning steps.  
To capture such dependencies, we adopt a graph-based formulation in which each trajectory $T_i$ is represented as a directed graph $G_i = (V_i, E_i)$. Nodes $V_i$ denote intermediate results, while edges $E_i$ encode logical dependencies or sub-questions that drive the reasoning forward. Each reasoning step $t_{ij}$ is thus represented as a pair $(v_{ij}, e_{ij})$, where $e_{ij}$ is the sub-question or logical operation and $v_{ij}$ is the resulting intermediate answer. As illustrated in~\cref{fig:intro_example}, darker sentences (Q) represent edges and lighter sentences (A) represent intermediate results, together forming a reasoning step.

This representation allows step selection to account not only for surface similarity but also for the topology of reasoning. In this setting, $Z$ is instantiated as a selected subgraph $G^* \subseteq G_i$.  
Then the objective of Eq.~\ref{obj:ib} becomes 
\begin{equation}
\label{obj:gib}
\min_{G^*} \; I(G_i; G^*) - \beta I(G^*; Y).
\end{equation}

\vspace{-3mm}
Since step-level labels are not observable, we approximate $Y$ by a consensus graph $G^{MC}$ aggregated from correct trajectories in the sampled set $\mathcal{S}$.  
For each reasoning graph $G_i$, we compute the Maximum Common Subgraph (MCS)~\citep{mccreesh2017partitioning} between $G_i$ and each correct graph $G_k \in \mathcal{G}_{\text{correct}}$. Then, each MCS highlights the reasoning components shared between $G_i$ and a correct solution.
Aggregating these pairwise MCS results yields $G^{MC}$, which serves as a consensus structure reflecting how $G_i$ aligns with correct reasoning. Details of MCS calculation and aggregation can be found in Appendix~\ref{app:algorithm}. Replacing $I(G^*;Y)$ with $I(G^*;G^{MC})$, the structure-alignment IB objective now becomes:
\begin{equation}
\label{obj:gib-mcs}
\min_{G^*} \; I(G_i; G^*) - \beta I(G^*; G^{MC}).
\end{equation}

\vspace{-5mm}
\paragraph{From IB to Soft-mask Relaxation.}
Directly solving the IB objective on graphs is intractable: selecting a discrete subgraph $G^* \subseteq G_i$ is a combinatorial problem, and estimating mutual information terms between discrete subgraphs and correctness signals is not feasible in practice.  
We therefore introduce a differentiable mask $\mathbf{p}_\theta = \{p_{\theta,ij}\}$ over steps, yielding a soft subgraph $G^* = G_i \odot \mathbf{p}$ as the compressed representation.  
For each step $t_{ij}$, the model predicts a selection probability $p_{ij} \in [0,1]$ that reflects its contribution to the retained reasoning structure.

\vspace{-2mm}
\paragraph{Approximating the IB Objective.}
The two mutual information terms in the IB objective are approximated with tractable surrogates:
% \begin{itemize}%[leftmargin=1em]
% \vspace{-2mm}
%     \item \emph{Compression.}  
    
\emph{Compression.} We start from the mutual information identity $I(G_i; G^*) = H(G^*) - H(G^* \mid G_i).$ 
To rigorously bound this objective, we adopt the Variational Information Bottleneck framework. 
Given that our step-selection mask consists of binary decisions (keep or drop), we naturally model the variational prior $r(z)$ as an independent Bernoulli distribution parameterized by a small constant $\epsilon < 0.5$.
Consequently, the compression objective is minimized via the KL divergence between the predicted mask $\mathbf{p}_\theta$ and this sparse prior:
\begin{equation}
\begin{aligned}
    & \mathcal{L}_{\text{compress}} \approx D_{KL}(\mathbf{p}_\theta \| r) \\
    & = \sum_{j} \left[ p_{\theta,ij} \log \frac{p_{\theta,ij}}{\epsilon} + (1-p_{\theta,ij})\log\frac{1-p_{\theta,ij}}{1-\epsilon} \right].
\end{aligned}
\end{equation}
Mathematically, this term decomposes into a sparsity penalty and a negative entropy component.
While some formulations of IB maximize entropy to learn a compressed stochastic representation, our goal is to select a single, determinate subgraph. 
By minimizing the entropy of the mask distribution (a component of the KL term), we force the model to make confident, binary-like decisions for each step (i.e., $p_{\theta,ij} \to 0$ or $1$). 
This directly encourages a compressed representation of the reasoning graph, thus satisfying the compression objective $I(G_i; G^*)$ in a variational manner.

% \vspace{-1mm}
\emph{Relevance.} To maximize the relevance term $I(G^*; Y)$, standard IB approaches typically require training an auxiliary classifier to estimate the conditional likelihood $p(Y|G^*)$, which is computationally expensive and unstable in black-box settings. Instead, we adopt a consensus-based surrogate strategy. We posit that the consensus structure $G^{MC}$, derived from the intersection of correct solutions, serves as a robust proxy for the essential reasoning information required to predict correctness ($Y=1$). Consequently, we approximate the maximization of mutual information by directly minimizing the Cross-Entropy (CE) between the predicted mask $p_\theta$ and the consensus mask $m_i$:$\mathcal{L}_{rel} = CE(p_\theta, m_i)$.

\vspace{-4mm}
\paragraph{Training Objective.}   
For each reasoning graph $G_i$, the model $f_\theta$ outputs a soft mask
$
\mathbf{p}_\theta = f_\theta(G_i),
$ 
where $p_{\theta,ij} \in [0,1]$ denotes the selection probability for step $t_{ij}=(v_{ij}, e_{ij})$.  
The consensus mask $\mathbf{m}_i$ is obtained by aligning $G_i$ with the maximum common subgraph $G^{MC}$ constructed from correct trajectories. Each $m_{ij} \in \{0,1\}$ indicates whether step $t_{ij}$ is part of the consensus reasoning structure.  
The final loss is
\begin{equation}
\label{eq:final-loss}
    \mathcal{L}(G_i) = H(\mathbf{p}_\theta) + \lambda \,\text{CE}(\mathbf{p}_\theta, \mathbf{m}_i).
\end{equation}

%\vspace{-4mm}
\paragraph{Inference.}  
At test time, given a new reasoning graph $G$ without any gold final-answer labels, the model outputs a probability mask $\mathbf{p}_\theta = f_\theta(G)$.  
For each step $t_{ij}=(v_{ij}, e_{ij})$, the step-wise confidence score is $c_{ij} = p_{\theta,ij}$.  
Steps with high probabilities are considered reliable, while those with low probabilities are flagged as logically inconsistent or error-prone.

\vspace{-2mm}
\subsection{Implementation}
%\vspace{-2mm}

Figure~\ref{fig:overview} illustrates the overall pipeline when we implement our SCA under the IB formulation. 

\textbf{(A) Structured Reasoning Trace from LLMs.} To model the reasoning process, we move beyond linear sequences to explicit reasoning graphs.
Specifically, we parse each trace into a reasoning graph $G_i=(V_i,E_i)$, where each step is represented as a node–edge pair $(v_{ij},e_{ij})$, where $v_{ij}$ is the result and $e_{ij}$ is the operation producing it~\citep{da2025understanding,chenteaching,amini2019mathqa,da2024llm}. For our experiments, we utilize LangFun-style prompting\footnote{\url{https://github.com/google/langfun}} to elicit these dependencies and extract graphs via a rule-based parser, though our framework is compatible with other parsing strategies (e.g., dependency parsing or implicit latent graphs). The full prompt is provided in Appendix~\ref{appedix:prompt}. A visualization is shown in Figure~\ref{fig:intro_example}.

\textbf{(B) Consensus Anchor Construction.} To obtain the compressed representation $Z$, we identify steps shared across correct solutions. For NIBS, we operationalize this alignment by comparing every step to steps in the correct set via a semantic similarity function. In GIBS, consensus anchors are identified by solving a maximum common subgraph (MCS) problem~\citep{mccreesh2017partitioning} between candidate and correct graphs. Steps that consistently appear in the MCS are treated as high-confidence. See Appendix~\ref{app:algorithm} for algorithms and Appendix~\ref{app:complexity} for complexity analysis. %Wrong trajectories are aligned against these anchors to highlight deviations.

\textbf{(C) IB-Based Confidence Attribution.} NIBS computes confidence from consensus anchors in closed form without training. GIBS learns a parameterized mapping using a BERT encoder for semantic step features and a 2-layer GCN for structural context; these representations are fused to predict soft selection probabilities $p_\theta$ via the IB objective (Eq.~\ref{eq:final-loss}). 
Note that while we adopt GCN based on preliminary empirical validation (see Appendix~\ref{app:sensitivity} for comparisons with other GNN architectures), these encoders remain replaceable components adaptable to different task complexities.

%\vspace{-1mm}
\section{Experiments}
%\vspace{-3mm}

\begin{table*}[t!]
\centering
\begin{adjustbox}{max width=\textwidth}
\begin{tabular}{ccccc|cccc|cccc}
\hline
LLM      & \multicolumn{4}{c}{\textbf{Llama3.1-8b}} & \multicolumn{4}{c}{\textbf{DeepSeek-R1-Distill-Qwen-32B}} & \multicolumn{4}{c}{\textbf{Phi4-reasoning}} \\ \cline{2-13} 
Metrics  & AUROC $\uparrow$       & AUCPR $\uparrow$      & ACC@80\% $\uparrow$    & ECE $\downarrow$    & AUROC $\uparrow$       & AUCPR $\uparrow$      & ACC@80\% $\uparrow$    & ECE $\downarrow$       & AUROC $\uparrow$       & AUCPR $\uparrow$      & ACC@80\% $\uparrow$    & ECE $\downarrow$   \\ \hline
\multicolumn{13}{c}{\cellcolor{gray!10}Dataset: GSM8K}                                                                                                   \\
P(true)  & 0.4016 & 0.4711 & 0.5283 & 0.5504 & 0.5159 & 0.5820 & 0.5840 & 0.5802 & 0.5251 & 0.6956 & 0.6934 & 0.6851 \\
SL(norm) & 0.4790 & 0.5240 & 0.5513 & \underline{0.2282} & 0.3700 & 0.5308 & 0.5262 & \textbf{0.1230} & 0.3851 & 0.5947 & 0.6799 & 0.2394 \\
Entropy  & 0.4105 & 0.4962 & 0.5141 & 0.2518 & 0.5203 & 0.5550 & 0.5235 & 0.2583 & 0.4623 & 0.6795 & 0.6664 & \underline{0.1648} \\
LECO     & 0.3862 & 0.4586 & 0.5319 & 0.3783 & 0.3202 & 0.4110 & 0.4883 & 0.3395 & 0.2885 & 0.5735 & 0.6344 & 0.4021 \\
\midrule
\rowcolor[HTML]{ECF4FF} 
Cos-Max  & 0.4537 & 0.5152 & 0.5513 & 0.3780 & 0.5269 & 0.5197 & 0.5676 & 0.3799 & 0.3494 & 0.5861 & 0.7010 & 0.1997 \\
\rowcolor[HTML]{ECF4FF} 
Cos-Mean & 0.6078 & 0.6211 & \underline{0.6175} & 0.3300 & 0.6633 & \underline{0.6933} & 0.5748 & 0.3323 & 0.5959 & 0.7703 & 0.7199 & \textbf{0.1556} \\
\rowcolor[HTML]{ECF4FF} 
NLI-Max  & \textbf{0.7096} & \textbf{0.7890} & 0.5908 & \textbf{0.1162} & \textbf{0.7450} & \textbf{0.6982} & \underline{0.6409} & \textbf{0.1456} & \underline{0.6600} & \underline{0.8141} & \underline{0.7446} & 0.2009 \\
\rowcolor[HTML]{ECF4FF} 
NLI-Mean & 0.5524 & 0.6103 & 0.5665 & 0.4376 & 0.6762 & 0.6508 & 0.6318 & 0.4189 & 0.5738 & 0.7704 & 0.7186 & 0.5527 \\  
% \midrule
\rowcolor[HTML]{DAE8FC} 
GIBS   & \underline{0.6910}       & \underline{0.7004}  & \textbf{0.6292} & 0.2293 & \underline{0.7289}            & 0.6712  & \textbf{0.6532} & 0.2867 & \textbf{0.7892} & \textbf{0.8172} & \textbf{0.8117} & 0.3354 \\
\midrule
\multicolumn{13}{c}{\cellcolor{gray!10}Dataset: MoreHopQA}  \\
P(true)  & 0.5228 & 0.5486 & 0.5450 & 0.5357 & 0.5177 & 0.6492 & 0.6606 & 0.6443 & 0.5086 & 0.6159 & 0.6362 & 0.6203 \\
SL(norm) & 0.4005 & 0.4506 & 0.5187 & 0.3168 & 0.2463 & 0.4894 & 0.6049 & 0.2829 & 0.3198 & 0.4966 & 0.5891 & 0.2751 \\
Entropy  & \underline{0.5510} & 0.5413 & \underline{0.5586} & \textbf{0.2148} & 0.6103 & 0.7150 & 0.6812 & \textbf{0.0952} & \underline{0.6012} & \underline{0.6702} & 0.5709 & \textbf{0.0520} \\
LECO     & 0.3280 & 0.4365 & 0.4921 & 0.3501 & 0.3116 & 0.5777 & 0.5642 & 0.3700 & 0.2760 & 0.4944 & 0.5782 & 0.4254 \\
\midrule
\rowcolor[HTML]{ECF4FF}
Cos-Max  & 0.3836 & 0.4446 & 0.5179 & 0.3687 & 0.3390 & 0.5129 & 0.6031 & 0.3066 & 0.3454 & 0.5284 & 0.6413 & 0.2614 \\
\rowcolor[HTML]{ECF4FF}
Cos-Mean & 0.5044 & 0.5487 & 0.5275 & 0.3160 & 0.5938 & 0.6919 & 0.6419 & 0.2434 & 0.4779 & 0.6492 & 0.6388 & \underline{0.2023} \\
\rowcolor[HTML]{ECF4FF}
NLI-Max  & 0.4937 & 0.5485 & 0.5347 & \underline{0.2159} & \underline{0.6663} & \underline{0.7766} & 0.6457 & \underline{0.1107} & 0.5801 & 0.6637 & \underline{0.6961} & 0.2323 \\
\rowcolor[HTML]{ECF4FF}
NLI-Mean & 0.5124 & \underline{0.5535} & 0.5314 & 0.3902 & 0.6291 & 0.6899 & \underline{0.6498} & 0.4305 & 0.5440 & 0.6428 & 0.6776 & 0.4610 \\
\rowcolor[HTML]{DAE8FC} 
GIBS   & \textbf{0.6471} & \textbf{0.6694} & \textbf{0.5602} & 0.3173 & \textbf{0.8084} & \textbf{0.8357} & \textbf{0.7051} & 0.1832 & \textbf{0.6619} & \textbf{0.6866} & \textbf{0.7053} & 0.3560 \\
\midrule
\multicolumn{13}{c}{\cellcolor{gray!10}Dataset: Math}                                                                                                    \\
P(true)  & 0.4584 & 0.4166 & 0.4551 & 0.4564 & 0.5055 & 0.6298 & 0.6226 & 0.6215 & 0.5277 & 0.7435 & 0.7554 & 0.7416 \\
SL(norm) & 0.5138 & 0.4627 & 0.4626 & 0.3250 & 0.4093 & 0.5433 & 0.6089 & 0.2319 & 0.3841 & 0.6597 & 0.7340 & 0.2220 \\
Entropy  &0.5120 & 0.4617 & 0.4550 & 0.2876 & 0.5190 & 0.6259 & 0.6225 & \underline{0.1814} & 0.5190 & 0.7351 & 0.7479 & 0.2023 \\
LECO     & 0.4378 & 0.4252 & 0.4382 & \underline{0.2229} & 0.3838 & 0.5836 & 0.5901 & \textbf{0.1813} & 0.4089 & 0.7279 & 0.7044 & 0.3957 \\
\midrule
\rowcolor[HTML]{ECF4FF}
Cos-Max  & 0.4313 & 0.4021 & 0.4513 & 0.4408 & 0.3376 & 0.5015 & 0.5937 & 0.3093 & 0.3845 & 0.6617 & 0.7656 & \underline{0.1568} \\
\rowcolor[HTML]{ECF4FF}
Cos-Mean & 0.5118 & 0.4848 & \underline{0.4607} & 0.3946 & 0.5363 & 0.6583 & 0.6177 & 0.2463 & \underline{0.6116} & \textbf{0.8503} & 0.7708 & \textbf{0.1013} \\
\rowcolor[HTML]{ECF4FF}
NLI-Max  & \underline{0.5173} & \textbf{0.5061} & \textbf{0.4743} & \textbf{0.1852} & 0.5310 & 0.6676 & 0.6340 & 0.2340 & 0.6043 & 0.8052 & \underline{0.7925} & 0.3090 \\
\rowcolor[HTML]{ECF4FF}
NLI-Mean & 0.5148 & 0.4881 & 0.4670 & 0.3356 & \underline{0.5407} & \underline{0.6694} & \underline{0.6267} & 0.4950 & 0.5739 & 0.8020 & 0.7712 & 0.5935 \\
\rowcolor[HTML]{DAE8FC} 
GIBS   & \textbf{0.5855}       & \underline{0.4890}      & 0.4513 & 0.2737     & \textbf{0.5806}            & \textbf{0.6831}           & \textbf{0.6359}     &0.3786        &  \textbf{0.6946} & \underline{0.8078}     & \textbf{0.8322}   & 0.4050  \\ \hline
\end{tabular}
\end{adjustbox}
%\vspace{-2mm}
\caption{\small Overall results for step-wise confidence attribution on the GSM8K, MoreHopQA, and Math datasets. Our proposed method, GIBS, consistently outperforms baseline methods, especially on more complex reasoning tasks. The \textbf{best} and \underline{second-best} results are highlighted. Higher AUROC, AUCPR, and ACC@80\% and lower ECE indicate better performance.}
\label{tab:main}
\vspace{-6mm}
\end{table*}

We conduct comprehensive experiments to evaluate the effectiveness of our SCA methods.\footnote{ Code can be found in \url{https://github.com/Xiao0o0o/stepwise-confidence-attribution}} Specifically, we address the following research questions:

\vspace{-1mm}
\noindent$\bullet$~\textbf{RQ1 (Accuracy)}: How accurately can our proposed methods (NIBS and GIBS) identify erroneous steps in a reasoning trace compared to strong baselines?

\vspace{-1mm}
\noindent$\bullet$~\textbf{RQ2 (Utility)}: Can the stepwise confidence scores from our methods be used to improve the final-answer accuracy of LLMs through targeted self-correction?

\vspace{-1mm}
\noindent$\bullet$~\textbf{RQ3 (Ablation \& Robustness)}: 
What is the contribution of each component in the GIBS framework?
How robust is our framework to variations in reasoning format, settings without final-answer labels, and domain shift?

%\vspace{-2mm}
\subsection{Experiment Settings}
\label{sec:settings}
%\vspace{-2mm}
Below we describe the experimental setup. Additional implementation details and sensitivity study are provided in Appendix~\ref{app:experiment-setting} and Appendix~\ref{app:sensitivity}.

\textbf{Datasets.}
We evaluate on three benchmarks for verifiable reasoning tasks, where final-answer correctness can be objectively determined: (1) GSM8K~\citep{cobbe2021gsm8k}, a widely used math word problem dataset; (2) Math~\citep{hendrycksmath2021math500}, competition-level problems requiring more complex reasoning; and (3) MoreHopQA~\citep{schnitzler2024morehopqa}, a multi-hop QA dataset testing generalization beyond math.

\vspace{-1mm}
\textbf{Compared Methods.}
We compare the following methods: (i) four \textit{white-box} CE approaches adapted to stepwise reasoning: SL(norm)~\citep{lin2024contextualized,cole2023selectively}, Token Entropy~\citep{kuhnsemantic}, P(true)~\citep{kadavath2022ptrue}, and LeCo~\citep{yao2024LECO}; (ii) our proposed NIBS family, which instantiates the closed-form IB solution with different similarity functions and aggregation strategies; and (iii) our GIBS model, which learns to align subgraphs with consensus structures. To our knowledge, there are no existing black-box methods that directly address stepwise confidence attribution. We adapt two methods from related tasks (reasoning evaluation~\citep{golovnevaroscoe} and error identification~\cite{mukherjeeparc}) for comparison. Results in Appendix~\ref{app:blackbox_baseline} show that our approach substantially outperforms these baselines.

\vspace{-1mm}
\textbf{Configurations.}
We evaluate three representative LLMs: LLaMA-3.1-8B-Instruct~\citep{grattafiori2024llama}, Phi-4-Reasoning~\citep{abdin2025phi4}, and DeepSeek-R1-Distill-Qwen-32B~\citep{guo2025deepseek}.
For each input, we sample $N=20$ traces with temperature $1.0$ to balance accuracy and diversity. 
For semantic similarity and equivalence, our framework is flexible in selecting a similarity function. We experiment with two approaches commonly used in UQ for LLMs: (1) cosine similarity between sentence embeddings from BERT~\citep{devlin2019bert}, and (2) semantic entailment predictions from an off-the-shelf NLI model~\citep{hedeberta,lingenerating}.
GIBS is trained on 2,000 reasoning graphs constructed from sampled solutions. At inference time, results for all methods are averaged over 10,000 trajectories per dataset.
For final-answer correctness, mathematical datasets are evaluated by exact match with gold answers, while QA datasets use GPT-4o as a judge. Additional details on model selection, hyperparameters, prompts, and evaluation setup are provided in Appendix~\ref{app:experiment-setting}.
\vspace{-1mm}

\textbf{Metrics.}
Following prior work in CE~\citep{davis2006aucpr,lingenerating,geifman2017accmetric}, we adopt four complementary metrics:
(i) \textbf{AUROC} and (ii) \textbf{AUCPR} evaluate ranking quality of confidence scores, with AUCPR being especially important under class imbalance since erroneous steps are sparse.
(iii) \textbf{ACC@80\%} measures selective prediction performance: we reject the 20\% of predictions with the lowest confidence and report accuracy on the remaining 80\%.
(iv) \textbf{ECE} (Expected Calibration Error) assesses calibration by comparing predicted confidence with empirical accuracy across bins.

\vspace{-2mm}
\subsection{Accuracy of Stepwise Confidence Attribution}
\vspace{-1mm}

\begin{figure}[t!]
    \centering
    \includegraphics[width=0.75\linewidth]{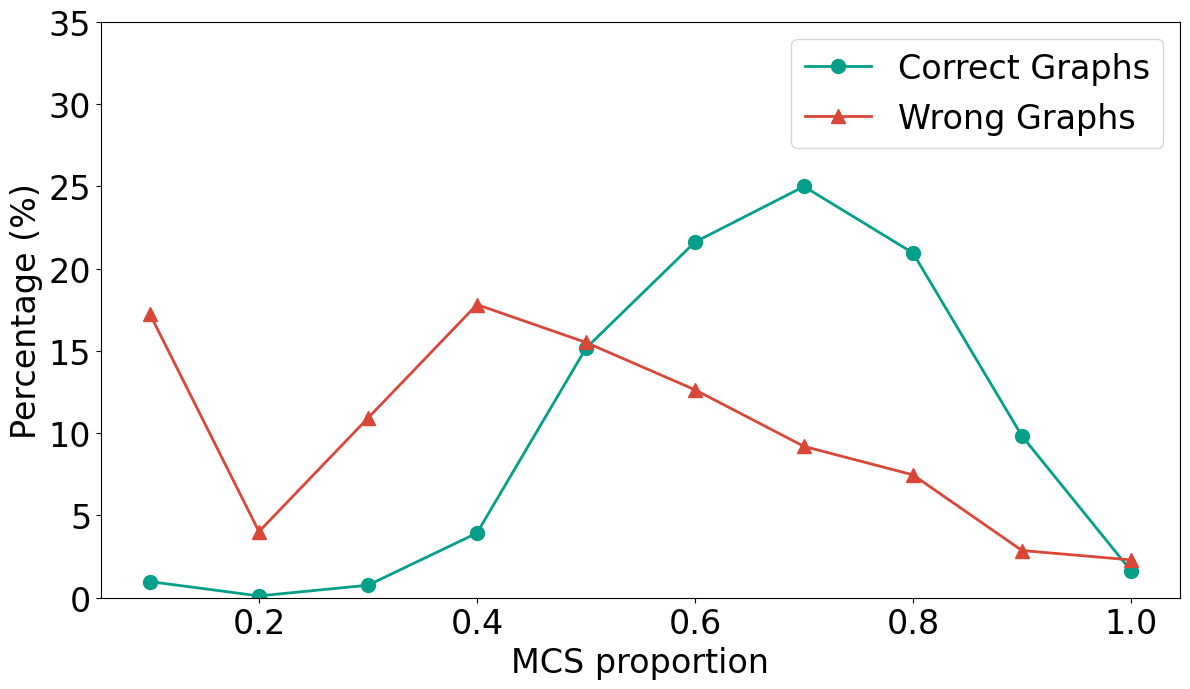}
    \vspace{-2mm}
    \caption{\small Distribution of average MCS proportion over 1,000 reasoning graphs. Correct graphs (green) concentrate near 0.8, while incorrect ones (red) peak near 0.4. This indicates that correct solutions are likely to follow a stable reasoning path.}
    \label{fig:mcs_size}
    \vspace{-5.5mm}
\end{figure}

We first provide empirical support for the intuition underlying our framework:  correct reasoning trajectories should exhibit stronger overlap in their common substructures.
Figure~\ref{fig:mcs_size} shows the distribution of average maximum common subgraph (MCS) scores across 1,000 reasoning graphs. Correct solutions concentrate around larger relative MCS sizes, while incorrect solutions peak around smaller values. This suggests that correctness is associated with stability and reproducibility of the reasoning path, while erroneous traces tend to diverge, producing fragmented structures. These observations provide empirical support for using consensus as a proxy signal for SCA.

We then compare NIBS and GIBS against strong white-box baselines for stepwise confidence attribution. Results in Table~\ref{tab:main} show consistent patterns across models and datasets.

$\bullet$~\emph{GIBS achieves the best AUROC in most settings.} Across three datasets and three LLMs, GIBS obtains the highest AUROC in 7 out of 9 configurations, demonstrating its strong ability to discriminate between correct and erroneous steps. This advantage stems from explicitly modeling structural dependencies: by aligning candidate subgraphs with consensus anchors, GIBS captures reasoning patterns that local similarity methods miss. The improvement is particularly pronounced on MoreHopQA, where reasoning spans multiple passages and logical alignment becomes essential. 

$\bullet$~\emph{NIBS provides competitive performance without training.} Although NIBS variants only capture semantic overlap without explicit structural modeling, they still achieve strong results across all datasets. Notably, the performance of these variants is robust to the choice of external models used for computing semantic similarity, with Cos-Mean and NLI-Max achieving similar results.
These results indicate that even non-parametric consensus alignment provides a reliable signal for stepwise confidence estimation.

$\bullet$~\emph{Our framework generalizes to human-annotated step-wise benchmarks.} We also evaluate on PRM800K~\citep{lightman2023prm800k}, a well-established stepwise reasoning benchmark containing pre-collected GPT-4 solutions with free-form chain-of-thought reasoning and high-quality human step-level annotations (+1, -1, 0 for each step). Unlike our main experiments, PRM800K requires no additional parsing; each reasoning trace is directly represented as a linear graph where sentences are sequentially connected. As detailed in Appendix~\ref{app:prm800k}, both NIBS and GIBS achieve strong performance on identifying incorrect steps, demonstrating the effectiveness of our approach on pre-collected traces with human-annotated ground truth.

% \vspace{-5mm}
\subsection{Self-Correction with Stepwise Confidence}
\vspace{-1mm}

\begin{figure}[t!]
    \centering
    \includegraphics[width=0.87\linewidth]{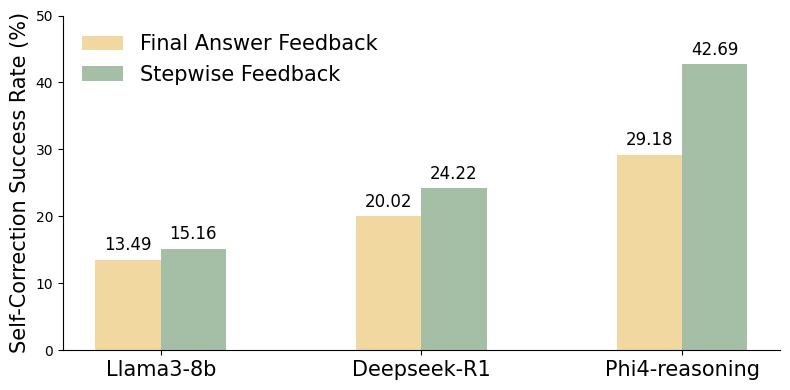}
    \vspace{-2mm}
    \caption{\small Effect of step-level feedback on correcting initially wrong answers in MoreHopQA. The baseline (yellow) provides only answer-level feedback, while our method (green) also highlights low-confidence steps.}
    \label{fig:acc_improvement}
    \vspace{-5mm}
\end{figure}

Previous experiments show we can detect errors; in this section, we show that this confidence attribution can actually guide effective self-correction. Specifically, we study whether highlighting low-confidence steps enables the model to successfully revise its erroneous answers.
Our evaluation considers MoreHopQA instances that the LLM initially answered incorrectly, and asks the LLM to regenerate its answer once, using two forms of feedback:
(1) \textbf{Final-answer feedback}: the model is only told that its final answer was wrong.
(2) \textbf{Stepwise feedback}: in addition to answer-level feedback, the model is shown its previous reasoning trace with step-level confidence scores.
We measure the self-correction success rate: the proportion of initially incorrect responses that the model successfully corrects after receiving feedback.
Figure~\ref{fig:acc_improvement} shows that stepwise feedback leads to substantially higher self-correction success rates than answer-level feedback alone. The improvement is most pronounced for stronger reasoning models such as DeepSeek-R1 and Phi-4-Reasoning, which can better exploit localized error signals to revise their reasoning.

We also present a case study in Appendix~\ref{app:case study} that illustrates how providing stepwise feedback enables an LLM to self-correct its errors. Since early mistakes can propagate and compound along the reasoning chain, we further report the performance of detecting the \emph{first} error in Appendix~\ref{app:first_error}.

\begin{table}[t!]
    \centering
    % \vspace{-5mm}
    \begin{tabular}{cccc}
    \toprule
    Method & GSM8K & MorehopQA & Math \\ \midrule
    GIBS                                                        & \textbf{0.7892}    &  \textbf{0.6619}     & \textbf{0.6946}         \\
    \begin{tabular}[c]{@{}c@{}}w/o Graph Encoder\end{tabular}   &  0.7228  &  0.6481  & 0.5961  \\
    \begin{tabular}[c]{@{}c@{}}w/o Edge Encoder\end{tabular} &    0.5186 &  0.3760    &   0.4763     \\
    \bottomrule
    \end{tabular}
    \caption{\small Performance of GIBS w.r.t. AUROC with and without the edge encoder or the graph encoder. }
    \label{tab:ablation_study}
    \vspace{-6mm}
\end{table}

\vspace{-2mm}
\subsection{Ablation Study}
\vspace{-1mm}

To investigate the effectiveness of different design choices, we conduct ablation experiments on Phi4-Reasoning across three datasets. The mask predictor in GIBS receives input from both an edge encoder, which captures local structural information, and a graph encoder, which provides global context. As shown in Table~\ref{tab:ablation_study}, removing either component leads to a noticeable drop in AUROC, confirming that both local and global signals are essential for accurate SCA.

\subsection{Generalization Analysis}

%\vspace{-2mm}
\paragraph{Generalization to Label-Free Settings} 
\label{sec:weak_supervision}
While our primary formulation utilizes final-answer correctness labels, practical deployment often lacks ground truth verification. We investigate whether our framework can generalize to such label-free scenarios by deriving consensus anchors from model-generated signals. We compare two unsupervised strategies: (1) \textit{All Trajectories}, which builds consensus from all sampled traces, and (2) \textit{Self-Consistency}, which replaces gold labels with majority-voted pseudo-correct trajectories.
As shown in Table~\ref{tab:weak_supervision}, using all trajectories suffers a clear performance drop due to noise from incorrect paths. The self-consistency variant, while still below the oracle \textit{Correct-only} setting, achieves competitive results on Deepseek and Phi4. We observe that performance correlates with pseudo-label quality: on Phi4, self-consistent trajectories overlap approximately $80\%$ with gold labels, resulting in only a minor performance drop; for Llama3.1-8B, overlap drops to about $28\%$, with corresponding degradation. These results indicate that our framework remains effective even without gold labels, as long as reasonably accurate reference trajectories can be obtained.

\begin{table}[t!]
    \centering
    \resizebox{\linewidth}{!}{%
    \begin{tabular}{lccc}
    \toprule
     & Llama3.1-8b & Deepseek & Phi4 \\
    \midrule
    All trajectories  & 0.5192 & 0.6479 & 0.5546 \\
    Self-consistency  & 0.5734 & 0.7843 & 0.6610 \\
    Correct-only      & \textbf{0.6471} & \textbf{0.8084} & \textbf{0.6619} \\
    \bottomrule
    \end{tabular}
    }
    \caption{\small Comparison of consensus anchor strategies. Correct-only performs best, while Self-consistency provides effective weak supervision with comparable results.}
    \label{tab:weak_supervision}
    \vspace{-5mm}
    \end{table}

\begin{figure}[t!]
    \centering
    \includegraphics[width=0.85\linewidth]{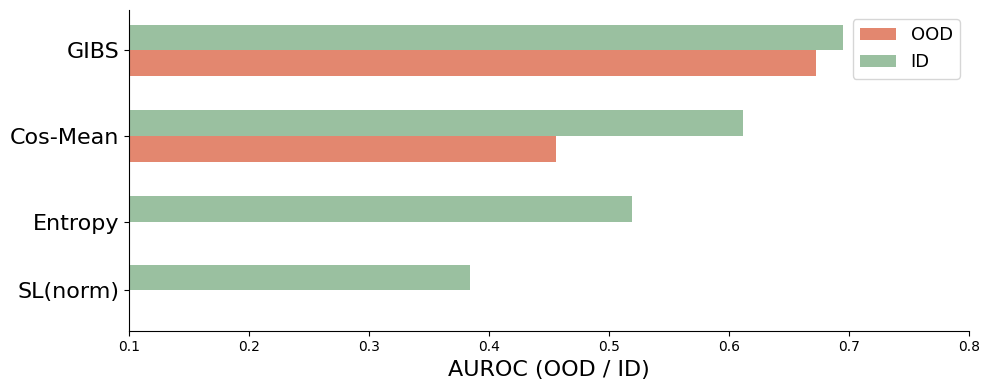}
    \vspace{-2mm}
    \caption{\small GIBS trained on MoreHopQA and tested on Math without re-training. GIBS consistently outperforms NIBS and white-box baselines under domain shift.}
    \label{fig:OOD}
    \vspace{-5mm}
\end{figure}

\vspace{-2mm}
\paragraph{Generalization to Out-of-distributions.}  
A practical CE method should remain effective beyond the domain it was trained on.  
To test this, we train GIBS with the trajectories generated by Phi4-Reasoning on MoreHopQA, a textual multi-hop QA dataset, as an in-distribution (ID) domain and evaluate it directly on the Math dataset without re-training as an out-of-distribution (OOD) domain. 
As shown in Figure~\ref{fig:OOD}, GIBS consistently achieves higher AUROC than NIBS and white-box baselines despite the domain shift.  
This advantage arises because NIBS relies on step similarity within the training domain, whereas GIBS learns structural representations under the IB principle, enabling it to capture abstract reasoning patterns that are less tied to domain-specific vocabulary.  
These results suggest that graph-based modeling provides strong robustness to domain variability.

%\vspace{-4mm}
\section{Conclusion}
%\vspace{-3mm}

This paper addresses stepwise confidence estimation in LLM reasoning, a key capability for diagnosing reasoning errors and enabling reliable model use. We introduced a framework for black-box LLMs based on the Information Bottleneck principle, with two complementary instantiations: NIBS, which assigns confidence via direct consensus alignment, and GIBS, which learns subgraph selection with consensus regularization. Across mathematical and textual reasoning benchmarks, both methods produced accurate, well-calibrated confidence scores that reliably localize erroneous steps. We further showed that stepwise confidence is actionable: integrating these signals into selective correction improved final-answer accuracy, and NIBS and GIBS demonstrated strong robustness and generalization.

\section*{Acknowledgment}
The work was partially supported by NSF award \#2442477 and \#2550203. Cheng is supported by the National Science Foundation (NSF) Grant \#2312862, NSF-Simons SkAI Institute, NSF CAREER \#2440542, NSF \#2533996, National Institutes of Health (NIH) \#R01AG091762, a Google Research Scholar Award, and Cisco gift grant. We thank Amazon Research Awards, Cisco Faculty Research Awards, and Toyota Faculty Research Awards. The authors acknowledge Google and OpenAI for providing us with API credits and Research Computing at Arizona State University for providing computing resources. 
The views and conclusions in this paper are those of the authors and should not be interpreted as representing any funding agencies.


% Authors are kindly asked to make their submissions as accessible as possible
% for everyone including people with disabilities and sensory or neurological
% differences. Tips of how to achieve this and what to pay attention to will be
% provided on the conference website \url{http://icml.cc/}.

% \section*{Software and Data}

% If a paper is accepted, we strongly encourage the publication of software and
% data with the camera-ready version of the paper whenever appropriate. This can
% be done by including a URL in the camera-ready copy. However, \textbf{do not}
% include URLs that reveal your institution or identity in your submission for
% review. Instead, provide an anonymous URL or upload the material as
% ``Supplementary Material'' into the OpenReview reviewing system. Note that
% reviewers are not required to look at this material when writing their review.

% \section*{Acknowledgements}

% \textbf{Do not} include acknowledgements in the initial version of the paper

% If a paper is accepted, the final camera-ready version can (and usually should)
% include acknowledgements.  Such acknowledgements should be placed at the end of
% the section, in an unnumbered section that does not count towards the paper
% page limit. Typically, this will include thanks to reviewers who gave useful
% comments, to colleagues who contributed to the ideas, and to funding agencies
% and corporate sponsors that provided financial support.

\section*{Impact Statement}
The expected impact of this work is primarily within the research community, by providing insights that may help improve model inference. While large language models (LLMs) may have a wide range of societal implications depending on their downstream use, this paper is intended to serve as an auditing aid for human oversight on LLMs rather than a replacement, which does not directly enable applications that raise novel ethical concerns.
There are many potential societal consequences of our work, none of which we feel must be specifically highlighted here. %This capability is crucial for safely deploying LLMs in reasoning-intensive domains where understanding where a model fails is as important as knowing that it failed. While our method enhances reliability, we acknowledge the potential risk of automation bias, where users might over-rely on visualized confidence scores; therefore, SCA is intended to serve as an auditing aid for human oversight rather than a replacement for it.

\nocite{langley00}
% \clearpage
\bibliography{example_paper}

@inproceedings{liu2023g,
  title={G-Eval: NLG Evaluation using Gpt-4 with Better Human Alignment},
  author={Liu, Yang and Iter, Dan and Xu, Yichong and Wang, Shuohang and Xu, Ruochen and Zhu, Chenguang},
  booktitle={Proceedings of the 2023 Conference on Empirical Methods in Natural Language Processing},
  pages={2511--2522},
  year={2023}
}

@article{augenstein2024factuality,
  title={Factuality challenges in the era of large language models and opportunities for fact-checking},
  author={Augenstein, Isabelle and Baldwin, Timothy and Cha, Meeyoung and Chakraborty, Tanmoy and Ciampaglia, Giovanni Luca and Corney, David and DiResta, Renee and Ferrara, Emilio and Hale, Scott and Halevy, Alon and others},
  journal={Nature Machine Intelligence},
  volume={6},
  number={8},
  pages={852--863},
  year={2024},
  publisher={Nature Publishing Group UK London}
}

@inproceedings{chen2026position,
  title={Position: Uncertainty Quantification in LLMs is Just Unsupervised Clustering},
  author={Chen, Tiejin and Da, Longchao and Liu, Xiaoouand Wei, Hua},
  booktitle={International Conference on Machine Learning},
  year={2026},
  organization={PMLR}
}

@article{gao2025llm,
  title={Llm-based nlg evaluation: Current status and challenges},
  author={Gao, Mingqi and Hu, Xinyu and Yin, Xunjian and Ruan, Jie and Pu, Xiao and Wan, Xiaojun},
  journal={Computational Linguistics},
  pages={1--27},
  year={2025},
  publisher={MIT Press 255 Main Street, 9th Floor, Cambridge, Massachusetts 02142, USA~…}
}

@article{wang2023aligning,
  title={Aligning large language models with human: A survey},
  author={Wang, Yufei and Zhong, Wanjun and Li, Liangyou and Mi, Fei and Zeng, Xingshan and Huang, Wenyong and Shang, Lifeng and Jiang, Xin and Liu, Qun},
  journal={arXiv preprint arXiv:2307.12966},
  year={2023}
}

@article{zhao2024auto,
  title={Auto-arena: Automating llm evaluations with agent peer battles and committee discussions},
  author={Zhao, Ruochen and Zhang, Wenxuan and Chia, Yew Ken and Xu, Weiwen and Zhao, Deli and Bing, Lidong},
  journal={arXiv preprint arXiv:2405.20267},
  year={2024}
}

@article{grattafiori2024llama,
  title={The llama 3 herd of models},
  author={Grattafiori, Aaron and Dubey, Abhimanyu and Jauhri, Abhinav and Pandey, Abhinav and Kadian, Abhishek and Al-Dahle, Ahmad and Letman, Aiesha and Mathur, Akhil and Schelten, Alan and Vaughan, Alex and others},
  journal={arXiv preprint arXiv:2407.21783},
  year={2024}
}

@article{schnitzler2024morehopqa,
  title={Morehopqa: More than multi-hop reasoning},
  author={Schnitzler, Julian and Ho, Xanh and Huang, Jiahao and Boudin, Florian and Sugawara, Saku and Aizawa, Akiko},
  journal={arXiv preprint arXiv:2406.13397},
  year={2024}
}

@article{da2025understanding,
  title={Understanding the uncertainty of llm explanations: A perspective based on reasoning topology},
  author={Da, Longchao and Liu, Xiaoou and Dai, Jiaxin and Cheng, Lu and Wang, Yaqing and Wei, Hua},
  journal={arXiv preprint arXiv:2502.17026},
  year={2025}
}

@article{da2024llm,
  title={Llm uncertainty quantification through directional entailment graph and claim level response augmentation},
  author={Da, Longchao and Chen, Tiejin and Cheng, Lu and Wei, Hua},
  journal={arXiv preprint arXiv:2407.00994},
  year={2024}
}

@inproceedings{devlin2019bert,
  title={Bert: Pre-training of deep bidirectional transformers for language understanding},
  author={Devlin, Jacob and Chang, Ming-Wei and Lee, Kenton and Toutanova, Kristina},
  booktitle={Proceedings of the 2019 conference of the North American chapter of the association for computational linguistics: human language technologies, volume 1 (long and short papers)},
  pages={4171--4186},
  year={2019}
}

@inproceedings{liu2025uncertainty,
  title={Uncertainty quantification and confidence calibration in large language models: A survey},
  author={Liu, Xiaoou and Chen, Tiejin and Da, Longchao and Chen, Chacha and Lin, Zhen and Wei, Hua},
  booktitle={Proceedings of the 31st ACM SIGKDD Conference on Knowledge Discovery and Data Mining V. 2},
  pages={6107--6117},
  year={2025}
}

@article{chen2025uncertainty,
  title={Uncertainty Quantification of Large Language Models through Multi-Dimensional Responses},
  author={Chen, Tiejin and Liu, Xiaoou and Da, Longchao and Chen, Jia and Papalexakis, Vagelis and Wei, Hua},
  journal={arXiv preprint arXiv:2502.16820},
  year={2025}
}

@article{chen2026every,
  title={Every Response Counts: Quantifying Uncertainty of LLM-based Multi-Agent Systems through Tensor Decomposition},
  author={Chen, Tiejin and Yao, Huaiyuan and Chen, Jia and Papalexakis, Evangelos E and Wei, Hua},
  journal={arXiv preprint arXiv:2604.08708},
  year={2026}
}

@article{patel2026llm,
  title={Are LLM Uncertainty and Correctness Encoded by the Same Features? A Functional Dissociation via Sparse Autoencoders},
  author={Patel, Het and Chen, Tiejin and Wei, Hua and Papalexakis, Evangelos E and Chen, Jia},
  journal={arXiv preprint arXiv:2604.19974},
  year={2026}
}

@article{lingenerating,
  title={Generating with Confidence: Uncertainty Quantification for Black-box Large Language Models},
  author={Lin, Zhen and Trivedi, Shubhendu and Sun, Jimeng},
  journal={Transactions on Machine Learning Research},
  year={2023}
}

@inproceedings{kuhnsemantic,
  title={Semantic Uncertainty: Linguistic Invariances for Uncertainty Estimation in Natural Language Generation},
  author={Kuhn, Lorenz and Gal, Yarin and Farquhar, Sebastian},
  booktitle={The Eleventh International Conference on Learning Representations},
  year={2023}
}

@inproceedings{besta2024graph,
  title={Graph of thoughts: Solving elaborate problems with large language models},
  author={Besta, Maciej and Blach, Nils and Kubicek, Ales and Gerstenberger, Robert and Podstawski, Michal and Gianinazzi, Lukas and Gajda, Joanna and Lehmann, Tomasz and Niewiadomski, Hubert and Nyczyk, Piotr and others},
  booktitle={Proceedings of the AAAI conference on artificial intelligence},
  volume={38},
  number={16},
  pages={17682--17690},
  year={2024}
}

@inproceedings{yeuncertainty,
  title={Uncertainty-Aware Step-wise Verification with Generative Reward Models},
  author={Ye, Zihuiwen and Melo, Luckeciano Carvalho and Kaddar, Younesse and Blunsom, Phil and Staton, Sam and Gal, Yarin},
  booktitle={ICLR Workshop: Quantify Uncertainty and Hallucination in Foundation Models: The Next Frontier in Reliable AI},
  year={2025}
}

@article{han2025mind,
  title={Mind the Generation Process: Fine-Grained Confidence Estimation During LLM Generation},
  author={Han, Jinyi and Li, Tingyun and Chen, Shisong and Shi, Jie and Wang, Xinyi and Yue, Guanglei and Liang, Jiaqing and Lin, Xin and Wen, Liqian and Chen, Zulong and others},
  journal={arXiv preprint arXiv:2508.12040},
  year={2025}
}

@article{zheng2024processbench,
  title={Processbench: Identifying process errors in mathematical reasoning},
  author={Zheng, Chujie and Zhang, Zhenru and Zhang, Beichen and Lin, Runji and Lu, Keming and Yu, Bowen and Liu, Dayiheng and Zhou, Jingren and Lin, Junyang},
  journal={arXiv preprint arXiv:2412.06559},
  year={2024}
}

@inproceedings{weng2023large,
  title={Large Language Models are Better Reasoners with Self-Verification},
  author={Weng, Yixuan and Zhu, Minjun and Xia, Fei and Li, Bin and He, Shizhu and Liu, Shengping and Sun, Bin and Liu, Kang and Zhao, Jun},
  booktitle={Findings of the Association for Computational Linguistics: EMNLP 2023},
  pages={2550--2575},
  year={2023}
}

@article{li2024llm-as-judge,
  title={Llms-as-judges: a comprehensive survey on llm-based evaluation methods},
  author={Li, Haitao and Dong, Qian and Chen, Junjie and Su, Huixue and Zhou, Yujia and Ai, Qingyao and Ye, Ziyi and Liu, Yiqun},
  journal={arXiv preprint arXiv:2412.05579},
  year={2024}
}

@article{jiao2025trustworthy,
  title={Trustworthy Reasoning: Evaluating and Enhancing Factual Accuracy in LLM Intermediate Thought Processes},
  author={Jiao, Rui and Zhang, Yue and Li, Jinku},
  journal={arXiv preprint arXiv:2507.22940},
  year={2025}
}

@article{lin2024contextualized,
  title={Contextualized sequence likelihood: Enhanced confidence scores for natural language generation},
  author={Lin, Zhen and Trivedi, Shubhendu and Sun, Jimeng},
  journal={arXiv preprint arXiv:2406.01806},
  year={2024}
}

@inproceedings{amini2019mathqa,
  title={MathQA: Towards Interpretable Math Word Problem Solving with Operation-Based Formalisms},
  author={Amini, Aida and Gabriel, Saadia and Lin, Shanchuan and Koncel-Kedziorski, Rik and Choi, Yejin and Hajishirzi, Hannaneh},
  booktitle={Proceedings of the 2019 Conference of the North American Chapter of the Association for Computational Linguistics: Human Language Technologies, Volume 1 (Long and Short Papers)},
  pages={2357--2367},
  year={2019}
}

@inproceedings{chenteaching,
  title={Teaching Large Language Models to Self-Debug},
  author={Chen, Xinyun and Lin, Maxwell and Sch{\"a}rli, Nathanael and Zhou, Denny},
  booktitle={The Twelfth International Conference on Learning Representations},
  year={2023}
}

@article{cobbe2021gsm8k,
  title={Training verifiers to solve math word problems},
  author={Cobbe, Karl and Kosaraju, Vineet and Bavarian, Mohammad and Chen, Mark and Jun, Heewoo and Kaiser, Lukasz and Plappert, Matthias and Tworek, Jerry and Hilton, Jacob and Nakano, Reiichiro and others},
  journal={arXiv preprint arXiv:2110.14168},
  year={2021}
}

@article{hendrycksmath2021math500,
  title={Measuring Mathematical Problem Solving With the MATH Dataset},
  author={Dan Hendrycks and Collin Burns and Saurav Kadavath and Akul Arora and Steven Basart and Eric Tang and Dawn Song and Jacob Steinhardt},
  journal={NeurIPS},
  year={2021}
}

@article{yao2024LECO,
  title={Learning from correctness without prompting makes LLM efficient reasoner},
  author={Yao, Yuxuan and Wu, Han and Guo, Zhijiang and Zhou, Biyan and Gao, Jiahui and Luo, Sichun and Hou, Hanxu and Fu, Xiaojin and Song, Linqi},
  journal={arXiv preprint arXiv:2403.19094},
  year={2024}
}

@inproceedings{cole2023selectively,
  title={Selectively Answering Ambiguous Questions},
  author={Cole, Jeremy and Zhang, Michael and Gillick, Dan and Eisenschlos, Julian and Dhingra, Bhuwan and Eisenstein, Jacob},
  booktitle={Proceedings of the 2023 Conference on Empirical Methods in Natural Language Processing},
  pages={530--543},
  year={2023}
}

@article{kadavath2022ptrue,
  title={Language models (mostly) know what they know},
  author={Kadavath, Saurav and Conerly, Tom and Askell, Amanda and Henighan, Tom and Drain, Dawn and Perez, Ethan and Schiefer, Nicholas and Hatfield-Dodds, Zac and DasSarma, Nova and Tran-Johnson, Eli and others},
  journal={arXiv preprint arXiv:2207.05221},
  year={2022}
}

@article{wei2022chain,
  title={Chain-of-thought prompting elicits reasoning in large language models},
  author={Wei, Jason and Wang, Xuezhi and Schuurmans, Dale and Bosma, Maarten and Xia, Fei and Chi, Ed and Le, Quoc V and Zhou, Denny and others},
  journal={Advances in neural information processing systems},
  volume={35},
  pages={24824--24837},
  year={2022}
}

@article{pandey2025adaptive,
  title={Adaptive graph of thoughts: Test-time adaptive reasoning unifying chain, tree, and graph structures},
  author={Pandey, Tushar and Ghukasyan, Ara and Goktas, Oktay and Radha, Santosh Kumar},
  journal={arXiv preprint arXiv:2502.05078},
  year={2025}
}

@misc{szymanski2024limitationsllmasajudgeapproachevaluating,
      title={Limitations of the LLM-as-a-Judge Approach for Evaluating LLM Outputs in Expert Knowledge Tasks}, 
      author={Annalisa Szymanski and Noah Ziems and Heather A. Eicher-Miller and Toby Jia-Jun Li and Meng Jiang and Ronald A. Metoyer},
      year={2024},
      eprint={2410.20266},
      archivePrefix={arXiv},
      primaryClass={cs.HC},
      url={https://arxiv.org/abs/2410.20266}, 
}

@misc{stechly2024selfverificationlimitationslargelanguage,
      title={On the Self-Verification Limitations of Large Language Models on Reasoning and Planning Tasks}, 
      author={Kaya Stechly and Karthik Valmeekam and Subbarao Kambhampati},
      year={2024},
      eprint={2402.08115},
      archivePrefix={arXiv},
      primaryClass={cs.AI},
      url={https://arxiv.org/abs/2402.08115}, 
}

@article{jacovi2024chain,
  title={A chain-of-thought is as strong as its weakest link: A benchmark for verifiers of reasoning chains},
  author={Jacovi, Alon and Bitton, Yonatan and Bohnet, Bernd and Herzig, Jonathan and Honovich, Or and Tseng, Michael and Collins, Michael and Aharoni, Roee and Geva, Mor},
  journal={arXiv preprint arXiv:2402.00559},
  year={2024}
}

@article{uesato2022solving,
  title={Solving math word problems with process-and outcome-based feedback},
  author={Uesato, Jonathan and Kushman, Nate and Kumar, Ramana and Song, Francis and Siegel, Noah and Wang, Lisa and Creswell, Antonia and Irving, Geoffrey and Higgins, Irina},
  journal={arXiv preprint arXiv:2211.14275},
  year={2022}
}

@article{zhang2024generative,
  title={Generative verifiers: Reward modeling as next-token prediction},
  author={Zhang, Lunjun and Hosseini, Arian and Bansal, Hritik and Kazemi, Mehran and Kumar, Aviral and Agarwal, Rishabh},
  journal={arXiv preprint arXiv:2408.15240},
  year={2024}
}

@article{wang2023math,
  title={Math-shepherd: Verify and reinforce llms step-by-step without human annotations},
  author={Wang, Peiyi and Li, Lei and Shao, Zhihong and Xu, RX and Dai, Damai and Li, Yifei and Chen, Deli and Wu, Yu and Sui, Zhifang},
  journal={arXiv preprint arXiv:2312.08935},
  year={2023}
}

@article{setlur2024rewarding,
  title={Rewarding progress: Scaling automated process verifiers for llm reasoning},
  author={Setlur, Amrith and Nagpal, Chirag and Fisch, Adam and Geng, Xinyang and Eisenstein, Jacob and Agarwal, Rishabh and Agarwal, Alekh and Berant, Jonathan and Kumar, Aviral},
  journal={arXiv preprint arXiv:2410.08146},
  year={2024}
}

@article{cao2023graphreason,
  title={Graphreason: Enhancing reasoning capabilities of large language models through a graph-based verification approach},
  author={Cao, Lang},
  journal={arXiv preprint arXiv:2308.09267},
  year={2023}
}

@article{fang2025graph,
  title={Graph of Verification: Structured Verification of LLM Reasoning with Directed Acyclic Graphs},
  author={Fang, Jiwei and Zhang, Bin and Wang, Changwei and Wan, Jin and Xu, Zhiwei},
  journal={arXiv preprint arXiv:2506.12509},
  year={2025}
}

@article{tyen2023llms,
  title={LLMs cannot find reasoning errors, but can correct them given the error location},
  author={Tyen, Gladys and Mansoor, Hassan and C{\u{a}}rbune, Victor and Chen, Peter and Mak, Tony},
  journal={arXiv preprint arXiv:2311.08516},
  year={2023}
}

@article{wang2022self,
  title={Self-consistency improves chain of thought reasoning in language models},
  author={Wang, Xuezhi and Wei, Jason and Schuurmans, Dale and Le, Quoc and Chi, Ed and Narang, Sharan and Chowdhery, Aakanksha and Zhou, Denny},
  journal={arXiv preprint arXiv:2203.11171},
  year={2022}
}

@inproceedings{mccreesh2017partitioning,
  title={A partitioning algorithm for maximum common subgraph problems},
  author={McCreesh, Ciaran and Prosser, Patrick and Trimble, James},
  booktitle={Proceedings of the 26th International Joint Conference on Artificial Intelligence},
  pages={712--719},
  year={2017}
}

@article{geifman2017accmetric,
  title={Selective classification for deep neural networks},
  author={Geifman, Yonatan and El-Yaniv, Ran},
  journal={Advances in neural information processing systems},
  volume={30},
  year={2017}
}

@inproceedings{davis2006aucpr,
  title={The relationship between Precision-Recall and ROC curves},
  author={Davis, Jesse and Goadrich, Mark},
  booktitle={Proceedings of the 23rd international conference on Machine learning},
  pages={233--240},
  year={2006}
}

@article{guo2025deepseek,
  title={Deepseek-r1: Incentivizing reasoning capability in llms via reinforcement learning},
  author={Guo, Daya and Yang, Dejian and Zhang, Haowei and Song, Junxiao and Zhang, Ruoyu and Xu, Runxin and Zhu, Qihao and Ma, Shirong and Wang, Peiyi and Bi, Xiao and others},
  journal={arXiv preprint arXiv:2501.12948},
  year={2025}
}

@article{abdin2025phi4,
  title={Phi-4-reasoning technical report},
  author={Abdin, Marah and Agarwal, Sahaj and Awadallah, Ahmed and Balachandran, Vidhisha and Behl, Harkirat and Chen, Lingjiao and de Rosa, Gustavo and Gunasekar, Suriya and Javaheripi, Mojan and Joshi, Neel and others},
  journal={arXiv preprint arXiv:2504.21318},
  year={2025}
}

@inproceedings{lightman2023prm800k,
  title={Let's verify step by step},
  author={Lightman, Hunter and Kosaraju, Vineet and Burda, Yuri and Edwards, Harrison and Baker, Bowen and Lee, Teddy and Leike, Jan and Schulman, John and Sutskever, Ilya and Cobbe, Karl},
  booktitle={The Twelfth International Conference on Learning Representations},
  year={2023}
}

@inproceedings{hedeberta,
  title={DEBERTA: DECODING-ENHANCED BERT WITH DISENTANGLED ATTENTION},
  author={He, Pengcheng and Liu, Xiaodong and Gao, Jianfeng and Chen, Weizhu},
  booktitle={International Conference on Learning Representations},
year = {2020}
}

@inproceedings{golovnevaroscoe,
  title={ROSCOE: A Suite of Metrics for Scoring Step-by-Step Reasoning},
  author={Golovneva, Olga and Chen, Moya Peng and Poff, Spencer and Corredor, Martin and Zettlemoyer, Luke and Fazel-Zarandi, Maryam and Celikyilmaz, Asli},
  booktitle={The Eleventh International Conference on Learning Representations},
year = {2022}
}

@inproceedings{mukherjeeparc,
  title={Premise-Augmented Reasoning Chains Improve Error Identification in Math reasoning with LLMs},
  author={Mukherjee, Sagnik and Chinta, Abhinav and Kim, Takyoung and Sharma, Tarun Anoop and Tur, Dilek Hakkani},
  booktitle={Forty-second International Conference on Machine Learning},
year = {2025}
}
\bibliographystyle{icml2026}

%%%%%%%%%%%%%%%%%%%%%%%%%%%%%%%%%%%%%%%%%%%%%%%%%%%%%%%%%%%%%%%%%%%%%%%%%%%%%%%
%%%%%%%%%%%%%%%%%%%%%%%%%%%%%%%%%%%%%%%%%%%%%%%%%%%%%%%%%%%%%%%%%%%%%%%%%%%%%%%
% APPENDIX
%%%%%%%%%%%%%%%%%%%%%%%%%%%%%%%%%%%%%%%%%%%%%%%%%%%%%%%%%%%%%%%%%%%%%%%%%%%%%%%
%%%%%%%%%%%%%%%%%%%%%%%%%%%%%%%%%%%%%%%%%%%%%%%%%%%%%%%%%%%%%%%%%%%%%%%%%%%%%%%
\newpage
\appendix
\onecolumn
\section{Notation Table}

\begin{table}[h]
\centering
\label{tab:notation}
\resizebox{0.9\linewidth}{!}{
\begin{tabular}{ll}
\toprule
\textbf{Symbol} & \textbf{Description} \\
\midrule
$x$ & Input problem or question \\
$y_i = (T_i, A_i)$ & $i$-th sampled output: reasoning trajectory $T_i$ and final answer $A_i$ \\
$T_i = \{t_{i1}, t_{i2}, \dots, t_{iL_i}\}$ & Reasoning trajectory with $L_i$ steps \\
$t_{ij} = (k_{ij}, a_{ij})$ & Step $j$ in trajectory $i$: description $k_{ij}$ and intermediate result $a_{ij}$ \\
$z_i \in \{0,1\}$ & Correctness label of final answer $A_i$ \\
$c_{ij} \in [0,1]$ & Confidence score assigned to step $t_{ij}$ \\
$\mathcal{S}$ & Sampled set of reasoning trajectories \\
$G_i = (V_i,E_i)$ & Graph representation of trajectory $T_i$ \\
$v_{ij}, e_{ij}$ & Node (intermediate result) and edge (reasoning operation) for step $t_{ij}$ \\
$G^{MC}$ & Consensus graph constructed from correct trajectories \\
$G^* \subseteq G_i$ & Subgraph selected by GIBS as compressed representation \\
$\mathbf{p}_\theta = \{p_{\theta,ij}\}$ & Soft selection mask predicted by the model \\
$\mathbf{m}_i = \{m_{ij}\}$ & Consensus mask indicating alignment with $G^{MC}$ \\
$X, Z, Y$ & Variables in IB formulation: input trajectory, compressed representation, and correctness \\
\bottomrule

\end{tabular}
}
\caption{Notation summary used throughout the paper. }
\end{table}

\section{Extended Problem Formulation and Discussion}
\label{app:problem_formulation}

In this section, we provide the formal definition of standard Answer-level Confidence Estimation (CE), which serves as the preliminary basis for our Stepwise framework.

Let an LLM be represented as a probabilistic model $\mathcal{M}$ that generates a response $y$ conditioned on input $x$. 
Answer-level CE aims to assign a reliability score to the final answer $A$. The definition of this score depends on the model's transparency:

\begin{itemize}
    \item \textbf{White-box Setting (Open-weights):} For models where internal states are accessible, confidence can be defined directly from token probabilities, e.g., $C(x,A) = p(A|x;\mathcal{M})$.
    \item \textbf{Black-box Setting (Closed-source):} For proprietary models where token-level probabilities are unavailable, confidence must be inferred from the agreement or consistency among $N$ sampled outputs $\{A_1, A_2, \dots, A_N\}$.
\end{itemize}

\begin{problem}[Answer-level CE]
Given an input $x$ and $N$ sampled answers $\{A_i\}_{i=1}^N$, the goal of answer-level CE is to learn a mapping
\[
f_{\text{ans}}: \{A_i\}_{i=1}^N \rightarrow \{c_i\}_{i=1}^N,
\]
where $c_i$ is the confidence score of answer $A_i$, estimated from observable signals such as self-consistency or semantic similarity among the sampled outputs.
\end{problem}

Our main paper extends this formulation to the granular level of reasoning steps (Problem~\ref{prob:SCA}), addressing the limitation that high answer-level confidence does not guarantee the correctness of individual reasoning steps.

\paragraph{Discussion: Internal Certainty vs. Correctness-Oriented Reliability}
A key motivation for our Stepwise Confidence Attribution (SCA) framework is the conceptual distinction between \textit{model certainty} and \textit{objective reliability}. 
In standard UQ, confidence is often equated with the model's internal likelihood (or negative entropy). However, in complex reasoning tasks, this proxy is often ill-calibrated due to two possible phenomena: 
(1) \textit{Confident Hallucinations:} A model may generate an incorrect reasoning step with high token probability if it follows a common misconception or linguistic pattern.
(2) \textit{The ``Uncommon but Correct'' Paradox:} A valid reasoning step might be creative or syntactically unique, resulting in low token probability. Relying solely on internal certainty would penalize such steps, treating valid variability as noise.

To address these limitations, we adopt a \textit{correctness-oriented} view, defining confidence as ``consistency with valid reasoning trajectories.'' 
One might ask: \textit{Why not simply measure consistency across all sampled trajectories, regardless of correctness?}
Our empirical analysis (and the logic of SCA) suggests that computing consensus over a mixed distribution of correct and incorrect traces introduces fundamental \textbf{ambiguity}. 
If the consensus metric (e.g., MCS size) is calculated against \textit{all} generated paths, the algorithm cannot distinguish whether a step aligns with a dominant \textit{correct} logic or a dominant \textit{error} mode (e.g., a common trap). 
As observed in our preliminary experiments, aligning with unfiltered trajectories causes the discrimination performance (AUROC) to drop significantly (often to near-random levels, $\sim 0.5$). 

Therefore, establishing anchors from \textbf{correct-only} trajectories is not merely an implementation detail but a theoretical necessity. It ensures that the ``confidence'' score specifically measures attribution to success, allowing creative but valid steps to receive high scores as long as they appear in at least one verified reasoning path.

\section{Algorithm Details}
\label{app:algorithm}

\subsection{NIBS Algorithm.} The procedure of NIBS is shown in Algorithm~\ref{alg:nibsc}.
Given a set of sampled trajectories, we first split them into correct and wrong subsets using the final-answer labels. 
For a target trajectory, we score each step by comparing it to steps drawn from correct trajectories: a step receives higher confidence if it has strong semantic matches (under the chosen similarity metric) to steps in correct solutions. 
Within-trajectory similarities are aggregated (e.g., by max or mean) to form the step’s score. 
This produces step-wise confidence without parameter learning and serves as a closed-form IB instantiation that compresses traces to consensus-supported steps.

\begin{algorithm}%[tb]
\caption{NIBS}
\label{alg:nibsc}
\begin{algorithmic}
   \STATE {\bfseries Input:} Sampled trajectories $\mathcal{S}=\{(T_i,A_i,z_i)\}_{i=1}^N$
   \STATE {\bfseries Output:} Stepwise confidence scores $\{c_{ij}\}$
   \STATE Partition $\mathcal{S}$ into $\mathcal{S}_{\text{correct}}=\{T_i\mid z_i=1\}$ and $\mathcal{S}_{\text{wrong}}$
   \FOR{each trajectory $T_i=\{t_{i1},\dots,t_{iL_i}\}$}
      \FOR{each step $t_{ij}$ in $T_i$}
         \STATE Compute similarities $\{\operatorname{sim}(t_{ij},t')\}$ for all steps $t'$ in trajectories from $\mathcal{S}_{\text{correct}}$
         \STATE Aggregate within-trajectory similarities by $\operatorname{Agg}(\cdot)$ (e.g., max or mean)
         \STATE Assign $c_{ij}$ by Eq.~\ref{eq:sim-CE}
      \ENDFOR
   \ENDFOR
   %\RETURN $\{c_{ij}\}$
   \STATE \textbf{Return} $\{c_{ij}\}$
\end{algorithmic}
\end{algorithm}

\subsection{GIBS Algorithm.} The training procedure of NIBS is shown in Algorithm~\ref{alg:gibsc-train}.
We convert each trajectory into a directed reasoning graph whose steps are represented as edge–node pairs. 
Using only correct graphs, we construct a consensus graph (e.g., via maximum common subgraph) and align it to each training graph to obtain a consensus mask on steps. In detail, the value for the consensus mask $\mathbf{m}_i$ can be computed as:

\[
m_{ij} \;=\; \frac{1}{|\mathcal{S}_{\text{correct}}|}\sum_{k}\mathbf{1}\!\big[v_{ij}\in \mathrm{MCS}(G_i, G_k)\big],
\]
where $\mathbf{1}[\cdot]$ is an indicator function.  
Thus, $m_{ij}=1$ if the step consistently appears in all MCS matches with correct graphs, $m_{ij}=0$ if it never appears, and intermediate values reflect partial support.

A trainable model outputs a soft selection mask over steps; the IB objective is approximated by the sum of a mask-entropy term (encouraging compression) and a cross-entropy term aligning the mask with the consensus anchors (encouraging relevance). 
%Optional budget and smoothness regularizers prevent trivial copying and promote coherent subgraphs. 
We optimize the model parameters by gradient descent to produce calibrated step-selection probabilities.

\begin{algorithm}%[tb]
\caption{GIBS: Training}
\label{alg:gibsc-train}
\begin{algorithmic}
   \STATE {\bfseries Input:} Reasoning graphs $\{G_i=(V_i,E_i)\}_{i=1}^N$ with labels $\{z_i\}$, model $f_\theta$
   \STATE {\bfseries Output:} Trained parameters $\theta$
   \STATE Partition $\{G_i\}$ into $\mathcal{G}_{\text{correct}}=\{G_i\mid z_i=1\}$ and $\mathcal{G}_{\text{wrong}}$
   \STATE Construct consensus graph $G^{MC}$ from $\mathcal{G}_{\text{correct}}$ (e.g., MCS aggregation)
   \FOR{each epoch}
      \FOR{each $G_i$}
         \STATE Represent each step as $t_{ij}=(v_{ij},e_{ij})$
         \STATE Align $G_i$ with $G^{MC}$ to obtain consensus mask $\mathbf{m}_i=\{m_{ij}\}$
         \STATE Compute soft mask $\mathbf{p}_\theta=f_\theta(G_i)$ with $p_{\theta,ij}\in[0,1]$
         \STATE Form soft subgraph $G_i^*=G_i\odot \mathbf{p}_\theta$
         \STATE Compute loss by Equation~\ref{eq:final-loss}
         %\STATE (optionally $+\ \alpha\,\Omega_{\text{budget}}+\mu\,\Omega_{\text{smooth}}$)
         \STATE Update $\theta$ by gradient descent
      \ENDFOR
   \ENDFOR
   \STATE \textbf{Return} $\theta$
\end{algorithmic}
\end{algorithm}

The inference procedure is shown in Algorithm~\ref{alg:gibsc-infer}. Given a new trajectory’s reasoning graph, the trained model predicts a soft mask over steps. 
Each step’s inclusion probability is reported as its confidence score. 
Optionally, the soft-masked subgraph can be visualized or passed to downstream routines (e.g., selective correction) to prioritize low-confidence steps while preserving logically consistent parts of the reasoning.

\begin{algorithm}%[tb]
\caption{GIBS: Inference}
\label{alg:gibsc-infer}
\begin{algorithmic}
   \STATE {\bfseries Input:} A new reasoning graph $G=(V,E)$, trained model $f_\theta$
   \STATE {\bfseries Output:} Step-wise confidence scores $\{c_{j}\}$
   \STATE Represent each step as $t_{j}=(v_{j},e_{j})$
   \STATE Compute soft mask $\mathbf{p}_\theta=f_\theta(G)$
   \FOR{each step $t_{j}$}
      \STATE Set confidence $c_{j}=p_{\theta,j}$
   \ENDFOR
   \STATE Optionally form $G^*=G\odot \mathbf{p}_\theta$ for visualization or downstream use
   \STATE {\bfseries Return} $\{c_{j}\}$
\end{algorithmic}
\end{algorithm}

\subsection{MCS Algorithm.}

MCS algorithm first identifies semantically similar edge pairs between the two reasoning graphs using an NLI model, and selects a small set of high-scoring candidates as seeds. 
Starting from each seed, it expands the common subgraph iteratively via a BFS, adding only those edges and nodes whose textual entailment scores exceed the thresholds and whose mappings remain consistent. In this paper, we set both thresholds $\tau_v,\tau_e = 0.7$.
Among all candidate expansions, the largest resulting subgraph is returned as the MCS along with the induced node and edge mappings. 
By integrating semantic similarity through entailment scores, the algorithm can align reasoning steps that are lexically different but semantically equivalent.

\begin{algorithm}%[tb]
\caption{FindMaximumCommonSubgraph}
\label{alg:mcs-nli}
\begin{algorithmic}
   \STATE {\bfseries Input:} Two reasoning graphs $G_1=(V_1,E_1), G_2=(V_2,E_2)$; reasoning texts $\mathcal{R}_1,\mathcal{R}_2$; thresholds $\tau_v,\tau_e$
   \STATE {\bfseries Output:} Common subgraph $G^{MC}$, node mapping $\pi_V$, edge mapping $\pi_E$
   \STATE Generate candidate edge pairs $\mathcal{C}=\{(e_1,e_2)\mid \text{NLI(Entail)}(\mathcal{R}_1(e_1),\mathcal{R}_2(e_2)) \ge \tau_e\}$
   \STATE Sort $\mathcal{C}$ by combined edge--node similarity, keep top-$K$ seeds
   \FOR{each seed pair $(e_1,e_2)\in \mathcal{C}$}
      \STATE Initialize subgraph $G^*$ with $(e_1,e_2)$, and mappings $\pi_V,\pi_E$
      \STATE Expand $G^*$ by BFS over neighbors of $e_1$:
      \STATE \quad For each $e'_1=(x_1,y_1)$, find best match $e'_2=(x_2,y_2)$ with entailment $\ge \tau_e$
      \STATE \quad If compatible with $\pi_V$, update $G^*$ and mappings
      \STATE Keep the largest $G^*$ found as the current best
   \ENDFOR
   \STATE {\bfseries return} $G^{MC}$ with similarity annotations, along with $\pi_V,\pi_E$
\end{algorithmic}
\end{algorithm}

\subsection{Complexity Analysis of the Heuristic MCS Algorithm}
\label{app:complexity}

We analyze the time and space complexity of the proposed MCS algorithm. Let $G_1 = (V_1, E_1)$ and $G_2 = (V_2, E_2)$ denote the two input reasoning graphs, and
write $m_1 = |E_1|$, $m_2 = |E_2|$, and $N = m_1 m_2$ for the number of edge pairs.
We denote by $\mathcal{T}_{\text{NLI}}$ the time required for a single forward pass of the
underlying NLI model, i.e., the cost of computing one entailment-based similarity score between
two text segments.

\paragraph{Time complexity.}
The proposed heuristic MCS algorithm proceeds in three phases.

\textbf{(1) Pairwise similarity estimation.}
In the first phase, the algorithm iterates over the Cartesian product
$E_1 \times E_2$ and, for each edge pair $(e_1, e_2)$, evaluates their semantic
compatibility. This involves a constant number of calls to the NLI model to score the correspondence between the edges and their incident nodes.
Thus the total cost of this phase is
\[
  O\bigl(N \cdot \mathcal{T}_{\text{NLI}}\bigr)
  \;=\;
  O\bigl(|E_1|\,|E_2| \cdot \mathcal{T}_{\text{NLI}}\bigr).
\]

\textbf{(2) Candidate ranking and pruning.}
Among all edge pairs that satisfy the similarity thresholds, the algorithm ranks
them according to their combined edge–node similarity and retains only the top-$K$
pairs (with $K=10$ in our experiments) as seed candidates.
If we denote by $M \leq N$ the number of pairs that pass the thresholds, then
sorting these candidates by score requires
\[
  O\bigl(M \log M\bigr) \;\subseteq\; O\bigl(N \log N\bigr)
  \;=\; O\bigl(|E_1|\,|E_2| \log(|E_1|\,|E_2|)\bigr)
\]
time in the worst case.
After sorting, truncating the list to the top-$K$ pairs is $O(1)$.

\textbf{(3) Heuristic subgraph expansion.}
Starting from each of the at most $K$ seed pairs, the algorithm performs a
BFS-style expansion on $G_1$, greedily adding neighboring edges and nodes as long as they admit compatible matches in $G_2$ within the pre-filtered candidate set.
For a fixed seed, this expansion touches at most $O(|V_1| + |E_1|)$ graph elements
of $G_1$, and since $K$ is a constant independent of the input sizes, the total cost of this phase is
\[
  O\bigl(K (|V_1| + |E_1|)\bigr)
  \;=\; O\bigl(|V_1| + |E_1|\bigr).
\]

Combining these contributions, the overall running time of the heuristic MCS algorithm can be bounded as
\[
  \mathcal{T}_{\text{total}}
  \;=\;
  O\bigl(|E_1|\,|E_2| \cdot \mathcal{T}_{\text{NLI}}
         + |E_1|\,|E_2| \log(|E_1|\,|E_2|)
         + |V_1| + |E_1|\bigr).
\]
In practice, the pairwise similarity phase is dominant, because it invokes
computationally expensive NLI forward passes for $O(|E_1|\,|E_2|)$ edge pairs.
Treating the NLI architecture and the hyperparameter $K$ as fixed, the overall complexity with respect to the graph sizes is thus essentially quadratic in the
number of edges:
\[
  \mathcal{T}_{\text{total}} = O\bigl(|E_1|\,|E_2| \cdot \mathcal{T}_{\text{NLI}}\bigr).
\]
\paragraph{Space complexity.}
The additional memory footprint is dominated by storing candidate edge pairs and
the intermediate MCS structures.
In the worst case, when many edge pairs pass the similarity thresholds, the
algorithm keeps $M = O(|E_1|\,|E_2|)$ candidates together with their scores.
The data structures used for BFS-based expansion are linear in the size of $G_1$,
i.e., $O(|V_1| + |E_1|)$, and are reused across different seeds.
Therefore, the overall auxiliary space complexity is
\[
  \mathcal{O}_{\text{space}}
  \;=\; O\bigl(M + |V_1| + |E_1|\bigr)
  \;\subseteq\; O\bigl(|E_1|\,|E_2|\bigr).
\]
The parameters and internal buffers of the NLI model contribute only a constant term with respect to the graph sizes, and do not affect the asymptotic behavior.

\section{Experiment Details}
\label{app:experiment-setting}

\subsection{Datasets} 
We evaluate our methods on three verifiable reasoning datasets spanning both mathematical and textual reasoning tasks:

(1) \textbf{GSM8K}~\cite{cobbe2021gsm8k}: A benchmark of grade-school math word problems, widely used for evaluating numerical reasoning and basic arithmetic logic.

(2) \textbf{Math}~\cite{hendrycksmath2021math500}: A more challenging dataset of competition-level mathematics problems. This dataset stresses the scalability of our methods to more complex mathematical reasoning.

(3) \textbf{MoreHopQA}~\cite{schnitzler2024morehopqa}: A multi-hop question answering dataset, where solving a query requires integrating evidence across multiple passages. MoreHopQA focuses on textual compositional reasoning and tests whether our framework generalizes to non-mathematical domains.

\subsection{Baseline}

We compare against four white-box CE baselines, each adapted to the step-wise reasoning setting:

(1) Normalized Sequence Likelihood (\textbf{SL(norm)})~\cite{lin2024contextualized,cole2023selectively}. For each reasoning step $t_{ij}$, we concatenate the description $k_{ij}$ and the intermediate result $a_{ij}$ together. The step-wise confidence is then defined as the SL(norm) of all tokens in this step:$
c_i^{SL} = \frac{1}{|T_i|}\sum_{t \in T_i}\operatorname{log}p_{\theta}(x_t | x_{<t}).
$

(2) Token-level Entropy (\textbf{Entropy})~\cite{kuhnsemantic}: Token Entropy computes the entropy of the predictive distribution at each token and then averages over tokens to obtain step-wise uncertainty, which is defined as
$
H_{\text{step}} \;=\; \frac{1}{|T|} \sum_{t=1}^T \sum_{k=1}^V p_{t,k} \log p_{t,k}.
$
Confidence is then taken as the $-H_{step}$.

(3) \textbf{P(true)}~\cite{kadavath2022ptrue}: Following prior work, we directly query the LLM itself to judge whether the current step is correct or not, and then use the probability assigned to the label \textit{true} as the confidence score. The prompt we used is in Appendix~\ref{appedix:prompt}.

(4) \textbf{LeCo}~\cite{yao2024LECO}: We also include LeCo, which introduces a logit-based confidence scoring method. It combines three components: the average token score, the step divergence score, and the inter-step transition score to compute an overall step confidence.

\subsection{Configurations}

For all experiments, we evaluate three representative LLMs: LLaMA-3.1-8B-Instruct, Phi-4-Reasoning, and DeepSeek-R1-Distill-Qwen-32B, and use identical prompting templates (Appendix~\ref{appedix:prompt}) to ensure fair comparisons. Decoding is performed with temperature set to $1.0$, and for each input question, we sample $N=20$ reasoning traces to balance the accuracy and diversity.

\textbf{Semantic Similarity.} To measure semantic similarity or equivalence, we experiment with two approaches commonly used in UQ for LLMs: (1) cosine similarity between sentence embeddings from \texttt{bert-base-uncased}~\citep{devlin2019bert}, and (2) semantic entailment predictions from the \texttt{DeBERTa-large-MNLI} model~\citep{hedeberta}. Our framework is flexible in this choice and can accommodate other similarity models.

\textbf{Reasoning Graph Extraction.} For GIBS, we extract reasoning graphs via structured generation: a LangFun-style template (Appendix~\ref{appedix:prompt}) asks the model to instantiate a Python \texttt{ReasoningGraph} class, and we extract nodes and edges with a simple rule-based parser. The edge- and node-level entailment thresholds $\tau_e, \tau_v$ used in the MCS procedure are both set to $0.7$, following prior work~\cite{da2025understanding,lingenerating}. A sensitivity analysis is also provided in Appendix~\ref{app:sensitivity}.

\textbf{GIBS Training.} We train GIBS on 2,000 reasoning graphs constructed from a mixture of correct and incorrect trajectories. Steps are encoded using embeddings from \texttt{bert-base-uncased}, and graph-level representations are obtained via a 2-layer GCN encoder with hidden dimension 128 and dropout 0.1. The model is optimized with Adam (learning rate $10^{-3}$) using early stopping based on validation loss.

\textbf{Evaluation.} Computing AUROC, AUCPR, and ACC@c\% requires step-level correctness labels (not confidence ground truth), which are obtained from GPT-4o using the evaluation templates in Appendix~\ref{appedix:prompt}. All reported results are averaged over 10,000 reasoning traces per dataset. Experiments are implemented in PyTorch and run on a single NVIDIA A100 GPU.

\subsection{Evaluation Metrics}

We evaluate how well the estimated confidence correlates with step-level correctness.
We adopt three complementary metrics to evaluate the effectiveness of our approach: AUROC, AUCPR, and ECE. 

(1) \textbf{AUROC} (Area Under the Receiver Operating Characteristic Curve)~\cite{lingenerating} measures the model’s ability to discriminate between correct and wrong steps across all thresholds, reflecting overall ranking performance. 

(2) \textbf{AUCPR} (Area Under the Precision–Recall Curve)~\cite{davis2006aucpr} focuses on ranking under class imbalance, which is crucial since erroneous steps typically form a small fraction of all steps. 

(3)\textbf{ACC@c\%}~\cite{geifman2017accmetric} is a standard metric in selective prediction, which reports the accuracy when retaining only the top-c\% most confident predictions. A reliable CE method will get a higher ACC@c\% by ranking correct predictions above incorrect ones.

(4) \textbf{ECE} (Expected Calibration Error) assesses the calibration of step-level confidence scores by comparing predicted confidence with empirical accuracy across bins, with lower values indicating better alignment.

\subsection{Prompt Template \& Few Shot Examples}
\label{appedix:prompt}

\begin{tcolorbox}[breakable,enhanced,
  colback=blue!3!white,
  colframe=blue!70!black,
  fonttitle=\bfseries,
  title={Prompt Template for Structured LLM Responses}]

\textbf{System Prompt:}  
Your role as an assistant involves thoroughly exploring questions through a systematic thinking process before providing the final, precise, and accurate solutions.

\medskip

\textbf{Prompt:}

{\small\color{black}\begingroup
\setlength{\parindent}{0pt}
\obeylines    
```
Please respond to the last INPUT\_OBJECT with OUTPUT\_OBJECT according to OUTPUT\_TYPE.\\
INSTRUCTIONS:
- Do NOT define or repeat any class or function.
- ONLY produce an OUTPUT\_OBJECT that instantiates the OUTPUT\_TYPE.
- The output must be valid Python using the given type names.
- Do NOT generate code, explanation, or helper variables.
- Only output an object like ReasoningGraph(...). \\

INPUT\_OBJECT:
  1 + 1 = \\
OUTPUT\_TYPE:
  Answer \\
  ```python
  class Answer:
    final\_answer: int
  ``` \\
OUTPUT\_OBJECT:
  ```python
  Answer(
    final\_answer=2
  )
  ``` \\
INPUT\_OBJECT:
{question} \\
OUTPUT\_TYPE:
ReasoningGraph \\
```python
class ReasoningNode:
  id: int
  description: str
  output: Union[int, float, str]
  depends\_on: list[int] \\

class ReasoningGraph:
  nodes: list[ReasoningNode]
  final\_answer: Union[int, float, str] \\
OUTPUT\_OBJECT:
"""
\endgroup}

\end{tcolorbox}

\begin{tcolorbox}[breakable,enhanced,
  colback=blue!3!white,
  colframe=blue!70!black,
  fonttitle=\bfseries,
  title={Regeneration Template Final Answer Feedback}]

\textbf{Structured LLM Responses Prompt + Prompt:}

{\small\color{black}\begingroup
\setlength{\parindent}{0pt}
\obeylines
\# NOTE: The previous final answer '\{\textcolor{blue}{answer}\}' is incorrect.\\
Please regenerate the OUTPUT\_OBJECT only.\\
Do not provide any reasoning, explanation, or extra text.\\
The OUTPUT\_TYPE remains ReasoningGraph.
\endgroup}

\end{tcolorbox}

\begin{tcolorbox}[breakable,enhanced,
  colback=blue!3!white,
  colframe=blue!70!black,
  fonttitle=\bfseries,
  title={Regeneration Template Stepwise Feedback}]

\textbf{Structured LLM Responses Prompt + Prompt:}

{\small\color{black}\begingroup
\setlength{\parindent}{0pt}
\obeylines
\# FEEDBACK:\\
The previous final answer '\{\textcolor{blue}{previous\_answer}\}' is incorrect.\\

\#\# Reasoning Process:\\
\{\textcolor{blue}{reasoning\_process}\}\\

\#\# Identified Errors (steps with low confidence):\\
\{\textcolor{blue}{error\_description}\}\\

Do not follow the identified error steps. Please reanalyze and regenerate a correct final answer that is different from the previous incorrect answer which is \{\textcolor{blue}{previous\_answer}\}. Return only the ReasoningGraph object without any explanations.
\endgroup}

\end{tcolorbox}

\begin{tcolorbox}[breakable,enhanced,
  colback=blue!3!white,
  colframe=blue!70!black,
  fonttitle=\bfseries,
  title={GPT Evaluation Prompt}]

\textbf{System Prompt:}  
You are evaluating a mathematical reasoning graph for correctness. Given the problem, correct answer, and a series of reasoning steps (edge-node pairs), determine if each step is mathematically and logically correct.

\medskip

\textbf{Prompt:}

{\small\color{black}\begingroup
\setlength{\parindent}{0pt}
\obeylines
Math Problem: \{\textcolor{blue}{question}\}
Correct Answer: \{\textcolor{blue}{answer}\}

Reasoning Graph (Edge-Node Pairs):
\{\textcolor{blue}{pairs\_text}\}

Instructions:
1. Evaluate each edge-node pair in the context of solving this problem
2. Consider if the edge description accurately describes what is being calculated
3. Check if the node value is mathematically correct given the edge description
4. Verify that each step logically follows from the problem or previous steps

For each pair, respond with:
- 1 if the edge-node pair is correct
- 0 if the edge-node pair is incorrect

Format your response as a comma-separated list of digits (no spaces), one digit per pair, without any explanation.
Example for 5 pairs: 1,0,1,1,0

Your evaluation:
\endgroup}

\end{tcolorbox}

\subsection{Parsing success rate}

We report the parsing success rate of structured reasoning traces in Table~\ref{tab:parsing-success}. Across datasets and models, the structured outputs can be parsed reliably. On GSM8K, all three models achieve over 99\% parsing success. On MoreHopQA and Math, the rates also remain high, with Llama3.1-8B achieving over 93\% across all datasets. These results indicate that modern instruction-tuned LLMs can reliably follow the structured output templates used in our framework.

\begin{table}[t]
\centering
\caption{Parsing success rates of structured reasoning traces across datasets and models.}
\label{tab:parsing-success}
\begin{tabular}{lccc}
\toprule
Model & GSM8K & MoreHopQA & Math \\
\midrule
Llama3.1-8B & 99.29\% & 96.11\% & 93.54\% \\
DeepSeek-R1-Distill-Qwen-32B & 99.35\% & 91.99\% & 99.17\% \\
Phi4-Reasoning & 99.69\% & 98.23\% & 97.92\% \\
\bottomrule
\end{tabular}
\end{table}

\subsection{Inference Efficiency}
\begin{table}
    \centering
    \begin{adjustbox}{max width=\columnwidth}
    \begin{tabular}{lcccccc}
    \toprule
    \multirow{2}{*}{Method} & \multicolumn{2}{c}{GSM8K} & \multicolumn{2}{c}{MoreHopQA} & \multicolumn{2}{c}{Math} \\
    \cmidrule(lr){2-3} \cmidrule(lr){4-5} \cmidrule(lr){6-7}
     & AUROC $\uparrow$  & Time $\downarrow$ & AUROC $\uparrow$  & Time $\downarrow$ & AUROC $\uparrow$  & Time $\downarrow$ \\
    \midrule
    GIB-based & \textbf{0.7892} & \textbf{0.05h} & \textbf{0.6619} & \textbf{0.02h} & 0.6946 & \textbf{0.005h} \\
    MCS-Only  & 0.7743 & 19h & 0.6593 & 10h & \textbf{0.7254} & 5h \\
    \bottomrule
    \end{tabular}
    \end{adjustbox}
    \caption{\small Comparison of GIB-based and MCS-Only methods w.r.t AUROC and inference time.}
    \label{tab:gib_mcs_results}
\end{table}

A key advantage of GIBS is its computational efficiency at inference time. While our training objective encourages the selected subgraph to align with the consensus among correct solutions, computing MCS explicitly is expensive and requires access to high-quality correct solution sets.

GIBS addresses this by training with soft consensus regularization, allowing the model to internalize consensus patterns. At inference, GIBS directly predicts step-level confidence via a forward pass, bypassing MCS computation entirely. As shown in Table~\ref{tab:gib_mcs_results}, GIBS achieves performance comparable to explicit MCS supervision while reducing inference time by over three orders of magnitude. This demonstrates that our approach effectively captures consensus structures without the computational overhead of exact MCS matching.

\section{Sensitivity Study}
\label{app:sensitivity}

We investigate the robustness of our framework with respect to two key
hyperparameters: (i) the NLI-based similarity thresholds used in MCS
construction, and (ii) the number of sampled trajectories $N$ used to form the
consensus.

\subsection{Sensitivity of NLI thresholds.}

In all main experiments, the entailment thresholds $\tau_e$ and $\tau_v$ are set to $0.7$, following prior work on semantic alignment~\citep{lingenerating,da2025understanding}. To assess sensitivity to this choice, we conduct a sensitivity analysis on MoreHopQA with Phi-4, jointly varying both thresholds in the range $[0.5, 0.9]$. As shown in Table~\ref{tab:nli-threshold}, the AUROC remains stable when $\tau_e, \tau_v$ lie in $[0.7, 0.8]$, and we observe noticeable degradation only when the thresholds are set too strictly ($> 0.8$) or too loosely ($< 0.7$). This supports our default choice $\tau_e, \tau_v = 0.7$ as a robust operating point.

\begin{table}[t]
  \centering
  \begin{tabular}{lccccc}
    \toprule
    Threshold & 0.5 & 0.6 & 0.7 & 0.8 & 0.9 \\
    \midrule
    AUROC     & 0.5821 & 0.6081 & 0.6619 & 0.6658 & 0.6112 \\
    \bottomrule
  \end{tabular}
  \caption{Sensitivity of AUROC to NLI thresholds on MoreHopQA with Phi-4.}
  \label{tab:nli-threshold}
\end{table}

\subsection{Sensitivity of the number of sampled trajectories.}

We also evaluate how performance depends on the number and diversity of sampled trajectories $N$ used for consensus construction. Table~\ref{tab:num-traj} reports AUROC on MoreHopQA with Phi-4 as we vary $N$ from $5$ to $25$. The results show that performance improves as $N$ increases and stabilizes once sufficient diversity is reached (approximately $N > 15$), indicating that the consensus becomes more reliable with richer sampling but remains robust beyond
this point.

In principle, black-box confidence estimation methods approximate the underlying output distribution by sampling multiple trajectories; hence, performance naturally scales with $N$ before plateauing, a behavior consistently observed in prior work~\citep{kuhnsemantic,lingenerating}. Following these works and our empirical analysis, we set $N = 20$ in all main experiments as a good trade-off between computational efficiency and stability.

\begin{table}[t]
  \centering
  \begin{tabular}{lccccc}
    \toprule
    \# Samples $N$ & 5 & 10 & 15 & 20 & 25 \\
    \midrule
    AUROC         & 0.5964 & 0.6024 & 0.6569 & 0.6619 & 0.6679 \\
    \bottomrule
  \end{tabular}
  \caption{Sensitivity of AUROC to the number of sampled trajectories $N$
  on MoreHopQA with Phi-4.}
  \label{tab:num-traj}
\end{table}

\subsection{Sensitivity of GNN backbones.}

Our learned method, GIBS, uses a graph neural network as a backbone to encode consensus-aware features. In the main experiments, we adopt a standard 2-layer GCN. To assess the role of the backbone architecture, we also experimented with GraphSAGE, GAT, and GIN on MoreHopQA with Phi-4 under the same training protocol (Table~\ref{tab:gnn-backbone}). We find that all backbones yield comparable behavior in this setting, and a simple GCN already provides a strong and reliable choice for modeling the required structural patterns.

\begin{table}[t]
  \centering
  \begin{tabular}{lcccc}
    \toprule
    GNN Backbone & GCN & GraphSAGE & GAT & GIN \\
    \midrule
    AUROC        & 0.6619 & 0.6446 & 0.5806 & 0.5815 \\
    \bottomrule
  \end{tabular}
  \caption{Influence of different GNN backbones on AUROC (MoreHopQA with Phi-4).}
  \label{tab:gnn-backbone}
\end{table}

\section{Additional Evaluation on Stepwise Correctness}

\subsection{Performance on PRM800K with Free-form Reasoning Traces} 
\label{app:prm800k}

We evaluate our method on PRM800K~\citep{lightman2023prm800k}, a well-established stepwise reasoning benchmark containing pre-collected GPT-4 solutions with free-form chain-of-thought reasoning and high-quality human step-level annotations. 

For each question, we sample $N=20$ diverse traces and use the gold step labels. NIBS is applied in the same way as in the main experiments. For GIBS, each CoT is represented as a linear graph: each sentence-level step is treated as an edge, edges are connected sequentially, and node contents are left empty. We train on 2,000 traces and evaluate on 10,000 traces. Since PRM800K consists of pre-generated outputs from closed-source LLMs, token-level logits are unavailable, and white-box baselines cannot be applied. Since PRM800K consists of outputs from the closed-source model, token-level logits are unavailable.

As shown in Table~\ref{tab:prm800k}, both NIBS and GIBS achieve strong performance across metrics, demonstrating that our framework can adapt effectively to free-form CoT. NIBS variants perform best overall, while GIBS remains competitive but less dominant than in structured settings, reflecting the limited explicit structure available for GIBS to exploit.

\begin{table}
    \centering
    \resizebox{0.5\linewidth}{!}{%
    \begin{tabular}{lccc}
    \toprule
    Method & AUROC~$\uparrow$ & AUCPR~$\uparrow$ & ACC@80\%~$\uparrow$ \\
    \midrule
    White-box methods & N/A    & N/A    & N/A    \\
    Cos-Max           & 0.6156 & 0.7840 & 0.7734 \\
    Cos-Mean          & 0.6821 & 0.8343 & 0.7860 \\
    NLI-Max           & 0.7666 & \textbf{0.9074} & 0.7899 \\
    NLI-Mean          & \textbf{0.8181} & 0.9019 & \textbf{0.8573} \\
    GIBS              & 0.6556 & 0.8203 & 0.7570 \\
    \bottomrule
    \end{tabular}
    }
    \caption{\small Performance of NIBS and GIBS on the PRM800K dataset with pre-collected GPT-4 traces. White-box methods require token probabilities, which are not applicable in this setting.}
    \vspace{-7mm}
    \label{tab:prm800k}
\end{table}

\subsection{Performance on Larger Models}

To evaluate whether SCA remains effective on larger models, we further conduct experiments with Qwen2.5-72B-Instruct on MoreHopQA. As shown in Table~\ref{tab:qwen72b}, GIBS consistently outperforms all baselines on both AUROC and AUCPR, achieving 0.7982 AUROC and 0.8385 AUCPR. These results indicate that the proposed graph-based stepwise attribution method remains effective when applied to stronger large-scale models.

\begin{table}[t]
\centering
\caption{Stepwise correctness attribution performance on MoreHopQA with Qwen2.5-72B-Instruct.}
\label{tab:qwen72b}
\begin{tabular}{lcc}
\toprule
Method & AUROC & AUCPR \\
\midrule
SL-norm & 0.4547 & 0.5218 \\
Entropy & 0.6114 & 0.6461 \\
LeCo & 0.5924 & 0.6973 \\
Cos-Max & 0.3136 & 0.5226 \\
Cos-Mean & 0.5722 & 0.6937 \\
NLI-Max & 0.6607 & 0.7637 \\
NLI-Mean & 0.6200 & 0.6835 \\
GIBS & \textbf{0.7982} & \textbf{0.8385} \\
\bottomrule
\end{tabular}
\end{table}

\subsection{False Positive Analysis of Low-Confidence Steps}

We further examine whether SCA incorrectly penalizes correct but uncommon reasoning steps. Specifically, we consider the bottom 20\% lowest-confidence steps flagged by GIBS and compute the fraction of them that are actually correct, which serves as the false positive rate of low-confidence attribution.

As shown in Table~\ref{tab:false-positive}, the false positive rates remain relatively low across datasets and models, ranging from 14.8\% to 21.6\%. This suggests that GIBS does not simply suppress uncommon reasoning patterns but more often assigns low confidence to genuinely problematic steps.

\begin{table}[t]
\centering
\caption{False positive rates of low-confidence steps identified by GIBS. We consider the bottom 20\% lowest-confidence steps and report the fraction of them that are actually correct.}
\label{tab:false-positive}
\begin{tabular}{lccc}
\toprule
Model & GSM8K & MoreHopQA & Math \\
\midrule
Llama3.1-8B & 15.9\% & 16.9\% & 21.6\% \\
Phi4-Reasoning & 15.4\% & 18.3\% & 20.2\% \\
DeepSeek-R1-Distill-Qwen-32B & 16.9\% & 14.8\% & 18.7\% \\
\bottomrule
\end{tabular}
\end{table}

\subsection{Human Validation of Stepwise Attribution}

To further validate the alignment between automatic attribution metrics and human judgment, we conduct a manual evaluation on MoreHopQA. For Phi4-Reasoning and DeepSeek-R1-Distill-Qwen-32B, we select 50 questions each and examine one incorrect trajectory per question. We then check whether the step with the lowest GIBS confidence matches the human-identified erroneous step.

For Phi4-Reasoning, whose AUROC is 0.66 in this setting, the lowest-confidence step matches the human-identified erroneous step in 60.0\% of cases (30/50). For DeepSeek-R1-Distill-Qwen-32B, whose AUROC is 0.81, the match rate increases to 80.0\% (40/50). This trend is consistent with the AUROC ranking, suggesting that our automatic metrics reasonably reflect the quality of step-level confidence attribution.

\section{Comparison with Adapted Black-Box Baselines}
\label{app:blackbox_baseline}

To our knowledge, there are no existing black-box methods that directly address stepwise confidence attribution for multi-step reasoning. However, several related approaches have been proposed for evaluating reasoning quality or detecting errors. We adapt two representative methods for comparison:

\textbf{ROSCOE}~\citep{golovnevaroscoe} restructures linear reasoning chains into Premise-Augmented Reasoning Chains by identifying premise links between steps, forming a directed acyclic graph. Originally designed to improve error identification by verifying each step under its premises, we adapt PARC by using its step-level verification scores as confidence scores.

\textbf{PARC}~\citep{mukherjeeparc} restructures linear reasoning chains into Premise-Augmented Reasoning Chains by identifying premise links between steps, forming a directed acyclic graph. Originally designed to improve error identification by verifying each step under its premises, we adapt PARC by using its step-level verification scores as confidence estimates.

As shown in Table~\ref{tab:blackbox_baseline}, both NIBS and GIBS substantially outperform these adapted baselines across all models and metrics. ROSCOE variants, designed for holistic reasoning evaluation rather than step-level error localization. PARC performs better by leveraging premise dependencies, but still falls short of our consensus-based methods, likely because PARC focuses on local premise-step verification rather than global consensus alignment. These results suggest that methods designed for related but distinct purposes do not transfer well to stepwise confidence attribution, highlighting the need for dedicated approaches like ours.

\begin{table}%[ht]
\centering
\resizebox{\linewidth}{!}{
\begin{tabular}{lccc ccc ccc}
\hline
 & \multicolumn{3}{c}{Llama3.1-8b} 
 & \multicolumn{3}{c}{Deepseek} 
 & \multicolumn{3}{c}{Phi4} \\
\cline{2-10}
 & AUROC & AUCPR & ACC@80\% 
 & AUROC & AUCPR & ACC@80\% 
 & AUROC & AUCPR & ACC@80\% \\
\hline
ROSCOE-ss & 0.477 & 0.517 & 0.503 & 0.408 & 0.584 & 0.528 & 0.421 & 0.603 & 0.542 \\
ROSCOE-sa & 0.524 & 0.565 & 0.541 & 0.492 & 0.600 & 0.612 & 0.488 & 0.594 & 0.613 \\
PARC      & 0.589 & 0.608 & 0.511 & 0.637 & 0.686 & 0.659 & 0.606 & 0.676 & 0.650 \\
NIBS      & 0.512 & 0.553 & 0.531 & 0.666 & 0.776 & 0.649 & 0.580 & 0.663 & 0.696 \\
GIBS      & \textbf{0.647} & \textbf{0.669} & \textbf{0.560} 
          & \textbf{0.808} & \textbf{0.835} & \textbf{0.705} 
          & \textbf{0.661} & \textbf{0.686} & \textbf{0.705} \\
\hline
\end{tabular}
}
\caption{Comparison with adapted black-box baselines on MoreHopQA. ROSCOE-ss and ROSCOE-sa are adapted from reasoning chain evaluation metrics; PARC is adapted from premise-based error identification. Both NIBS and GIBS substantially outperform these baselines.}
\label{tab:blackbox_baseline}
\end{table}

\section{Case Study} 
\label{app:case study}
\begin{figure}
    \centering
    \includegraphics[width=0.95\linewidth]{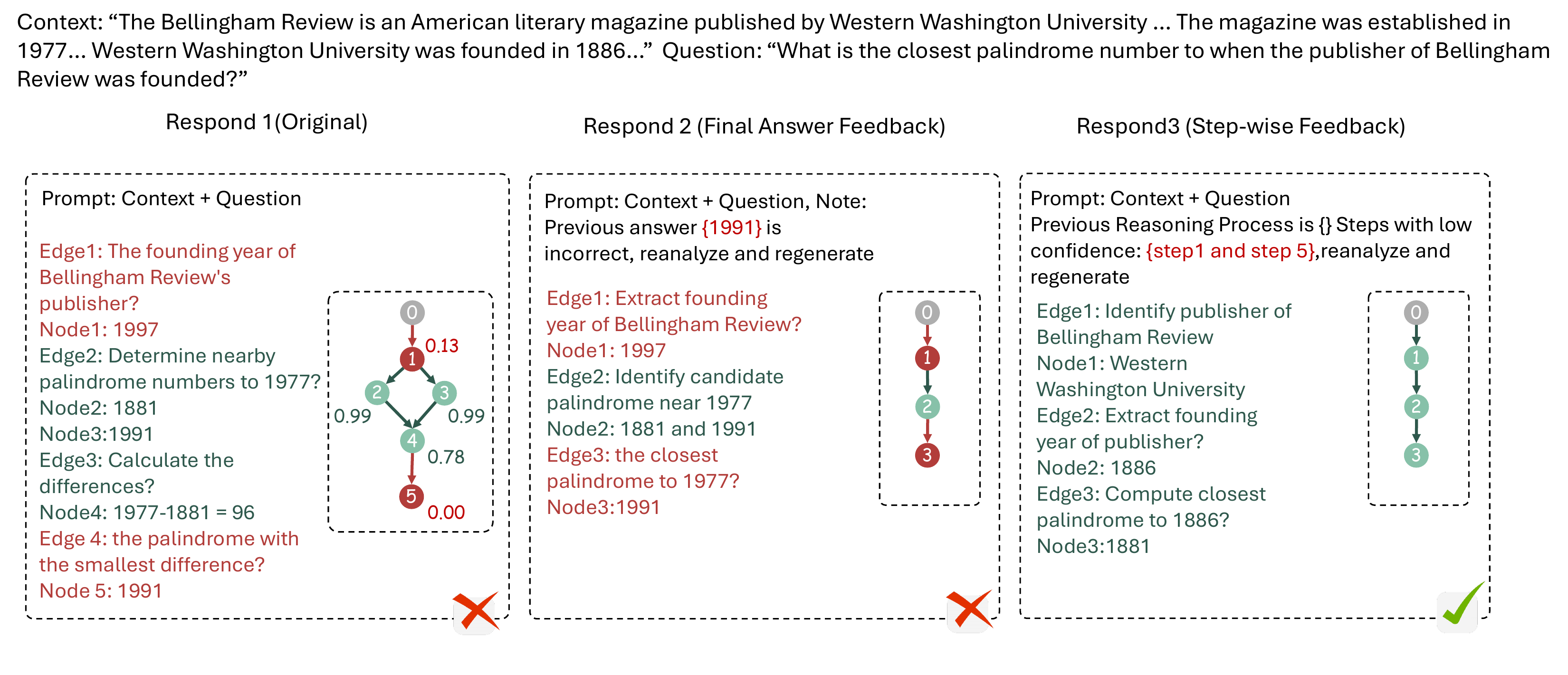}
    \caption{\small  A case study on the MoreHopQA dataset comparing the effect of different feedback types. Providing targeted, step-wise feedback on low-confidence reasoning steps is effective at guiding the model to correct its root error, whereas providing simple final-answer feedback is not.}
    \label{fig:case_study}
\end{figure}

Figure~\ref{fig:case_study} provides a case study from the MoreHopQA dataset that qualitatively demonstrates the advantage of our method. Initially, the LLM produces an incorrect answer by focusing on the wrong entity (the magazine's founding year instead of the publisher's). Our method correctly identifies this flawed premise by assigning low confidence to the initial reasoning steps. While simple final-answer feedback is insufficient for the model to find this root error, our step-wise feedback explicitly flags the low-confidence steps. This targeted guidance prompts the model to reconsider its flawed premise, revise its reasoning trajectory, and arrive at the correct answer. This case illustrates that LLMs can effectively self-correct when their reasoning uncertainty is accurately localized.

\section{First-Error Step Detection}
\label{app:first_error}

A natural concern is that step-wise confidence attribution might be dominated by \emph{later} erroneous steps: once a trajectory has already gone off course, subsequent steps often become trivially inconsistent with the correct reasoning pattern. Thus, for debugging and
test-time correction, identifying the \emph{first} wrong step is most critical, since all downstream errors are typically propagated from this point.

Our framework inherently assigns confidence scores to \emph{all} intermediate steps, so it also produces a score for the earliest erroneous step in each trace. To disentangle performance on the true failure point from that on subsequent steps, we conduct an additional evaluation that focuses exclusively on the first error. Concretely, for each trajectory that contains at least one incorrect step, we locate the first step whose correctness label is $0$ and keep \emph{only} this step when computing the metrics; all later steps in that trajectory are excluded from the evaluation set.

Table~\ref{tab:first-error-step} summarizes the results on MoreHopQA for three LLMs. Across all models, GIBS achieves the best performance, indicating that it is particularly effective at ranking the first erroneous step above the correct ones.

\begin{table}[ht]
\centering
\begin{tabular}{lccc}
\hline
 & Llama3.1-8b & Deepseek & Phi4 \\
\cline{2-4}
 & AUROC & AUROC & AUROC \\
\hline
P(true)    & 0.5126 & 0.5152 & 0.5090 \\
SL(norm)   & 0.4941 & 0.3009 & 0.4724 \\
Entropy    & 0.5570 & 0.6004 & 0.6837 \\
LECO       & 0.3839 & 0.3307 & 0.3884 \\
Cos-Max    & 0.4275 & 0.4529 & 0.4772 \\
Cos-Mean   & 0.4863 & 0.5313 & 0.5099 \\
NLI-Max    & 0.5375 & 0.7110 & 0.6632 \\
NLI-Mean   & 0.5285 & 0.6544 & 0.5824 \\
GIBS       & \textbf{0.6164} & \textbf{0.7885} & \textbf{0.6841} \\
\hline
\end{tabular}
\caption{First-error step detection performance (AUROC) on MoreHopQA dataset.}
\label{tab:first-error-step}
\end{table}

\end{document}